# Forma mentis networks predict creativity ratings of short texts via interpretable artificial intelligence in human and GPT-simulated raters

Edith Haim[1], Natalie Fischer[1], Salvatore Citraro[2], Giulio Rossetti[2] and Massimo Stella[1]

[1]*CogNosco Lab, Department of Psychology and Cognitive Science, University of Trento, Rovereto, Trento*

[2] *Institute of Information Science and Technologies "A. Faedo" (ISTI), National Research Council (CNR), Pisa*

**Author Note**

| | |
|---|---|
| Edith Haim | https://orcid.org/0009-0000-4244-904X |
| Natalie Fischer | https://orcid.org/0009-0002-2991-612X |
| Salvatore Citraro | https://orcid.org/0000-0002-5021-4790 |
| Giulio Rossetti | https://orcid.org/0000-0003-3373-1240 |
| Massimo Stella | https://orcid.org/0000-0003-1810-9699 |

**Corresponding author**

Edith Haim - email address: edith.haim@unitn.it

## Abstract

Creativity is a fundamental skill of human cognition. We use textual forma mentis networks (TFMN) to extract network (semantic/syntactic associations) and emotional features from approximately one thousand human- and GPT3.5-generated stories. Using Explainable Artificial Intelligence (XAI) we test whether features relative to Mednick's associative theory of creativity can explain creativity ratings assigned by humans and GPT-3.5. Using XGBoost, we examine 3 scenarios: (i) human ratings of human stories, (ii) GPT-3.5 ratings of human stories, and (iii) GPT-3.5 ratings of GPT-generated stories. Our findings reveal that GPT-3.5 ratings differ significantly from human ratings not only in terms of correlations but also because of feature patterns identified with XAI methods. GPT-3.5 favours "its own" stories and rates human stories differently from humans. Feature importance analysis with SHAP scores shows that: (i) network features are more predictive for human creativity ratings but also for GPT-3.5´s ratings of human stories; (ii) emotional features played a greater role than semantic/syntactic network structure in GPT-3.5 rating its own stories. These quantitative results underscore key limitations in GPT-3.5´s ability to align with human assessments of creativity. We emphasise the need for caution when using GPT-3.5 to assess and generate creative content, as it does not yet capture the nuanced complexity that characterises human creativity.

# Forma mentis networks predict creativity ratings of short texts via interpretable artificial intelligence in human and GPT-simulated raters

## 1. Introduction

Creativity is a fundamental aspect of human knowledge. Understanding how conceptual associations convey creativity has become a focal point of scientific research. In a cognitive context, creativity can be defined as the ability to generate ideas or products that are both novel and valuable (Boden, 1998; Runco & Chand, 1995). When it comes to texts, creativity involves the use of language in innovative ways, encompassing originality in word choice, syntactic structure, and the overall thematic expression (Weinstein et al., 2022; Johnson et al., 2023). Evaluating the creativity of texts presents a challenge due to the subjective nature of creativity and the specific characteristics of textual data. Metrics such as word infrequency, unique word combinations, syntax uniqueness, rhyme, and phonetic similarity have been proposed as good explainers for variance in human creativity ratings (Weinstein et al., 2022). Creativity in texts generated by large language models (LLMs), however, has not yet received as much attention, which leaves a gap in the literature.

In short narratives, creativity has been identified through various stylistic and qualitative features. These include new word formations, unexpected turns in the plot, and original and vivid settings (D'Souza, 2021). An unexpected turn in the plot refers to a surprising development or twist that deviates from the anticipated storyline. Vivid settings, on the other hand, are detailed and imaginative descriptions of the environment in which the story takes place, allowing readers to visualise and immerse themselves in the narrative (D'Souza, 2021). Beyond these qualitative features, the creativity level of a short narrative can also be assessed on a more quantitative level. Psychometric approaches have been used to evaluate creativity in texts (Johnson et al., 2023). Understanding and assigning scores for creativity in short narratives has benefitted from network science and machine learning approaches. Although these approaches are powerful in combining human and Artificial Intelligence (AI) early models for ratings, these methods face challenges, such as the difficulty in creating standardised measures that accurately reflect all the nuanced and multifaceted aspects of creative writing (Weinstein et al., 2022). Nevertheless, previous research has underscored significant correlations between the structure of semantic networks and creativity levels in individuals, suggesting that network features may serve as reliable indicators of narrative creativity (Guo et al., 2019; Heinen & Johnson, 2018; Semeraro et al., 2024). Furthermore, recent studies have leveraged semantic network features and machine learning algorithms to reduce subjectivity in creativity ratings and automatically assign more objective creativity scores (Acar, 2023; Beaty & Johnson, 2021; Heinen & Johnson, 2018; Johnson et al., 2023).

**1.1 Quantitative assessments of semantic and syntactic features of creative stories**

Advancements in computational models, such as Bidirectional Encoder Representations from Transformers (BERT), have demonstrated significant predictive power in assessing creativity based on the structural components of narratives (Johnson et al., 2023). BERT is a deep learning language model that understands the context of words in a text by analysing their co-occurrences (Vaswani et al., 2017). To represent the structural components of narratives, BERT uses word embeddings, i.e. numerical representations of words that capture meanings, syntactic properties, and relationships with other words. These embeddings place words in a continuous vector space, where words with similar meanings are located closer to each other. This allows BERT to effectively understand and process the context of words in a text, enhancing its ability to analyse and generate human-like language (Johnson et al., 2023). Whereas embeddings represent words as numerical vectors of non-interpretable word-word co-occurrences in texts, there are also other approaches to quantifying the structure of associative knowledge representing creativity levels in texts. For instance, the field of cognitive network science (Siew et al., 2019) promotes interpretable models where associative knowledge is represented as a set of conceptual relationships linked by one or several types of associations (Stella et al., 2024). In this way, a semantic network is a representational model of associations between concepts, and the network structure enables quantitative measurements apparently unavailable to unstructured texts. For instance, there are syntactic relationships between words in texts, which, although not immediately visible, are crucial for sentence meaning. These syntactic relationships help in understanding how words function together within a sentence and which roles each of these words play, thereby contributing to the overall meaning (Haim & Stella, 2023). Consider, for instance, the changed meaning in the following sentences when rearranging the word order and, thus, rearranging the syntactic dependencies: "Peter gave a flower to the cow. The cow gave a flower to Peter." By analysing the amount and type of syntactic relationships, we can reconstruct semantic frames or attribute semantic richness (i.e. node degree) to a given concept. This allows for a deeper understanding of how ideas are connected and how complex meanings are constructed in a text (Haim et al., 2024).

Textual forma mentis networks (TFMNs) provide a powerful tool for this kind of analysis (Stella, 2020). TFMNs automatically extract the structure of mindsets - the layout of associations connecting concepts with each other - from textual data. These cognitive networks automatically extract conceptual associations and sentiment labels from texts (Semeraro et al., 2024). The resulting network structure is informative of the cognitive layout of conceptual associations, representing how authors organised and associated their knowledge around various topics. Unlike "black box" machine learning approaches, TFMNs provide transparent insights into the semantic content and emotional context of the analysed text, including emotional scores expressing how rich in emotional words a given text is compared to a random model. For instance, TFMNs have been successfully used to reconstruct issues such as the gender gap in science, revealing how knowledge is organised and perceived around this topic (Stella, 2020).

TFMNs provide semantic network features for the analysed texts by using the EmoAtlas library in Python. EmoAtlas leverages lexicons that have been validated in psychological studies to detect the eight basic emotions identified in Plutchik's theory of basic emotions (Plutchik, 1980). Plutchik's theory posits that there are eight primary emotions, each of which has a polar opposite: joy and sadness, trust and disgust, fear and anger, and surprise and anticipation. These emotions can combine to form more complex feelings and play a crucial role in how humans experience and express emotions (Plutchik, 1980). These emotional features are quantified by calculating z-scores, which determine how frequently emotion words appear in the text compared to a null model (Semeraro et al., 2024). This approach identifies emotions that are significantly over- or underrepresented, and is explained in more detail in Section 2.3.

Semeraro et al. (2024) found that both BERT and TFMNs display different features when reproducing creativity ratings for the same textual data. While BERT focuses on semantic aspects of creativity, TFMNs highlight syntactic and emotional relationships, with each method offering unique insights into the creative process of narratives (Semeraro et al., 2024).

The semantic network features gained from a complex network, such as network distance, clustering coefficient or degree centrality, can be used as structural components to assess the creativity level of short stories (Semeraro et al., 2024). Network distance, i.e. average shortest path length (ASPL), refers to the average number of steps, hops or connections required to link two concepts within the network (Haim & Stella, 2023). From a cognitive perspective, distance captures the relatedness of concepts, indicating how many syntactic specifications or relationships can separate two concepts in a given text. Higher distances between entries suggest more remote associations, which are often linked to higher creativity levels (Haim et al., 2024), as they imply a broader range of conceptual connections. The clustering coefficient, on the other hand, reflects how frequently triads of concepts are interconnected, potentially mirroring syntactic structures such as subject-verb-object triplets (Haim & Stella, 2023). High clustering coefficients indicate tightly knit groups of concepts, which can imply a well-structured narrative. Conversely, lower clustering coefficients might suggest more diverse and loosely connected ideas, contributing to creative storytelling by avoiding overly predictable or conventional associations. Furthermore, degree centrality measures the number of direct connections a node has with other nodes in the network (Siew et al., 2019). In a cognitive context, high degree centrality signifies that certain pivotal concepts are highly connected to other concepts within the narrative. This centrality can highlight key ideas or themes that serve as focal points in the story, contributing to its depth and complexity (Kenett, 2019).

The network approach can importantly fuel ideas coming from cognitive psychology, like the Associative Theory of Creativity proposed by Mednick (1962). This psychological theory posits that creativity arises from the ability to form connections between distant and seemingly unrelated concepts within a semantic network. According to this theory, the more distant the connections, the higher the creativity. Thus, assuming that creativity interacts with a structured system where one can define a notion of distance between concepts, network distance can be employed as a proxy for testing the presence of creativity levels, with greater distances suggesting higher levels of creative potential (Heinen & Johnson, 2018; Johnson et al., 2023).

**1.2 Main aims of this work**

Importantly, automated assessments of creativity levels in stories can be applied also to the investigation of machine psychology. Considering Large Language Models (LLMs) as cognitive agents able to search for concepts in a certain knowledge space and assemble narratives, one can consider the task of assessing creativity features in LLMs as similar to the task of assessing creativity in humans and their stories. However, to the best of our knowledge, despite some studies investigating creativity in LLMs directly via psychometric questionnaires (Góes et al., 2023), assessing LLMs´ creativity features from the text that they produced is apparently an unexplored research direction. This gap can be addressed with the above frameworks from cognitive network science.

The current work lies at the intersection of machine learning, creativity research and cognitive network science, providing a quantitative comparison between the features that characterise those stories being rated as creative by either humans or GPT-3.5. Our approach leverages Explainable Artificial Intelligence (XAI) in order to address the limitations of subjective human evaluations (Beaty & Johnson, 2021; Heinen & Johnson, 2018). Furthermore, by contrasting not only ratings but also human and GPT-3.5-generated stories, we seek to investigate the shared or diverging attributes of creativity as perceived by both human raters and AI.

The paper at hand builds on the foundational work by Johnson et al. (2023) and extends the understanding of how quantitative cognitive network features, beyond distance, can predict creativity ratings. We investigate the influence of 13 quantitative network and emotional features of texts on creativity scores. These additional measures offer deeper insights into the organisation and thematic consistency of narratives, potentially correlating with their creative potential (Berahmand et al., 2018; Orwig et al., 2021; Siew, 2013).

Even before the launch and widespread availability of ChatGPT, researchers have debated the creative potential of AI, but few studies have quantified it in direct contrast to human creativity. In our work, we utilise the human dataset provided online by Johnson et al. (2023) and compare it to a newly generated dataset with an equivalent number of stories ("participants") created by GPT-3.5. By analysing the TFMNs and the emotional features of stories produced by GPT-3.5, we aim to understand how these AI-generated narratives are structured and subsequently compare these structures to creativity ratings also provided by GPT-3.5. This dual analysis allows us to assess not only the structural elements of the stories but also the features GPT-3.5 considers important when rating the creativity of a narrative. It is relevant to mention that at the current time of writing up this manuscript, GPT-3.5 is not available any more for use via the web interface but it remains available in the GPT store and for usage via API, thus still enabling scientific experiments. In the following subsection we outline the ideas behind emotional features of stories.

**1.3 Beyond semantics: Incorporating emotions in short stories´ creativity ratings**

Besides syntactic and semantic structure, the role of emotions in narrative creativity should not be ignored. Emotions profoundly influence memory, cognitive processing, imagination, and the overall structure of narratives (Hustvedt, 2011; Miall, 2011). Research by Vrana et al. (2019) has demonstrated that the coherence of narratives varies with emotional content, with neutral narratives exhibiting greater coherence than those recounting traumatic events. By incorporating emotion analysis into our assessment framework, we aim to explore how the presence and intensity of basic emotions (anger, trust, surprise, disgust, joy, sadness, fear, and anticipation) influence creativity ratings. We implement this emotion analysis using the EmoAtlas library (Semeraro et al., 2024), which captures emotions through psychologically validated datasets and provides interpretable measures.

**1.4 Manuscript outline**

This work is structured around four key objectives, aimed at deepening our understanding of the interplay between narrative structure, emotion, and perceived creativity. Firstly, we focus on predicting creativity levels based on specific network and emotion features of the stories. The network features we try to leverage include concept centrality (e.g. degree, PageRank) and features of network structure (e.g. average shortest path length, clustering coefficient, diameter) with some levels of interpretability within the literature of cognitive network science. By analysing these features, we aim to identify which elements are most closely related to high or low creativity ratings given by human or GPT-3.5 raters. We also investigate the differences across four raters to determine if any rater assigns more importance to certain features over others, reflecting individual biases in creativity assessment.

Secondly, we compare the semantic network features and emotion scores between stories generated by human participants and those produced by GPT-3.5. This comparison will reveal differences in how these narratives are structured and the emotional content they convey.

Thirdly, we have also tasked GPT-3.5 with rating the original human stories collected by Johnson et al. (2023), thus obtaining a third set of ratings of GPT-3.5 rating human stories. We use this as an additional point of comparison to assess how GPT-3.5 rates stories generated by GPT-3.5 differently from human stories and how the GPT-3.5 ratings differ from the human ratings. We expect that certain network features, such as a higher clustering coefficient and degree centrality, will correlate strongly with higher creativity ratings, reflecting well-defined themes and pivotal narrative elements. On an emotional level, we anticipate that narratives with higher z-scores for positive emotions like joy and anticipation will be rated as more creative, consistent with findings that positive emotional valence enhances creative flexibility, which is a critical component of creative thinking. For instance, positive emotional states can reduce cognitive switch costs, indicating enhanced cognitive flexibility and better performance in creative problem-solving tasks (Amabile et al., 2005; Wang et al., 2017).

Finally, we also hypothesise that differences will emerge in the creativity assessments between human raters and GPT-3.5 raters, with AI potentially focusing more on structural coherence and complexity due to its training on vast amounts of text data (Boden, 1998). By addressing these questions, our project aims to contribute to the understanding of narrative creativity and the potential of AI in creative assessments.

## 2. Methodology

### 2.1 Materials

*2.1.1 Human dataset*

The human dataset used in this study was generated as part of Study 2 of a series on divergent semantic integration (DSI) by Johnson et al. (2023). In the original study, Johnson et al. (2023) recruited 153 participants using Amazon's Mechanical Turk and compensated them $5.00 for their participation. The demographic distribution was as follows: Mean age = 38.62 (range: 22-70 years); 82 women, 68 men, 3 non-binary individuals; 97% English first-language speakers; 78% White, 9% African-American, 3% Asian-American, and 9% other.

Participants were instructed to write a total of 7 creative short stories (excluding one practice story). Three-word prompts were provided to them and all 3 words had to be included in the respective story. Each story was to be 4-6 sentences long and written within 4 minutes. The instruction script that was provided to the participants in the original study can be found in Table 1 in direct comparison with the modified instruction script we used for replicating the process with GPT-3.5.

After going through the instructions, the participants were given the following prompts, one by one, in a randomised order per participant: stamp-letter-send; gloom-payment-exist;

organ-empire-comply; statement-stealth-detect; belief-faith-sing; petrol-diesel-pump; year-week-embark.

The stories were then rated by four human raters. Each rater was supposed to rate all 1071 stories but some ratings were missing for some stories in the dataset available from Johnson et al. (2023). For the purpose of our study, we cleaned the dataset by keeping only the stories that were rated by all four raters, discarding those with missing ratings and thus ensuring usability and reliability in our results. We discarded 74 of the originally 1071 stories. This left us with 997 stories to investigate.

*2.1.2 GPT dataset*

To replicate the original study as closely as possible, we started out by testing and adapting the original instruction script. We did this by interacting directly with the online ChatGPT interface of GPT-3.5 and performing prompt engineering in an interactive way, testing responses directly on the web interface. When we were getting adequate and consistent responses to our prompts we settled on a script, which can be found in Table 1 in direct comparison with the original script. We used the same three-word prompts as Johnson et al. (2023). We then manually generated the stories using the GPT-3.5 web interface, resetting the anonymous chat after each "individual". We did this to minimise the risk of previous response influences between distinct participants. We retrieved a total of 1071 stories generated by GPT-3.5: 7 from each of the 153 "participants". There were a few instances in which the "regenerate" function had to be employed, primarily due to technical errors or token limitations imposed by OpenAI, the company behind ChatGPT. In those cases, no story had been generated for the participant or prompt in question.

| **Original Script** (Johnson et al., 2023) | **Modified Script** |
|---|---|
| **Instruction Script**<br>Please read these instructions carefully as we will ask you questions about them later. In the next task, you are required to write a very short story. You will be given 3 words, and you must write a story that is 4 to 6 sentences long, that includes all 3 words. Try to use your imagination and be creative when writing your story. You have 4 minutes to write your story. Once you are finished, click the arrow button. When the 4 minutes are up, the task will move on.<br><br>**Instruction Check Questions**<br>What should you type in response to the three words?<br>You will write one sentence for each creative story.<br><br>**Practice Trial**<br>pencil-paper-write<br>*(Johnson et al., 2023)* | **Instruction Script**<br>Please read these instructions carefully as we will ask you comprehension questions about them before you can begin the task. You are required to write seven very short stories. You will be given 3 words, and you must write a story that is 4 to 6 sentences long and that includes all 3 words. Try to use your imagination and be creative when writing your story. After you have completed the story, the next three words will be given to you and you will write a new distinct story. You will also need to write one practice story before starting the actual task. Are you ready to receive the comprehension questions?<br><br>**Instruction Check Questions**<br>What is expected of you as a response to the three words?<br>You will write one sentence for each creative story: true or false?<br><br>**Practice Trial**<br>Now please do a practice run with the following three words: pencil - paper - write<br><br>**Main task**<br>You're ready to begin the main task now.<br>Prompt 1 - Your three words are: …<br>Prompt 2 - Your three words are: …<br>[…] |

*Table 1. Instruction scripts for prompting GPT-3.5 with creating creative short stories.*

While the core content of our script remained unchanged, we had to make some adjustments in wording for best results. The most significant change was the addition of the word "distinct", as the AI model initially did not generate a new story for each prompt, instead continuing the previously written one. We also excluded the time limit from the instructions, as it was not relevant in this context: aside from longer waiting times during high-demand periods, GPT-3.5 typically required only seconds to write a story.

**2.2 Creativity Ratings of Stories**

*2.2.1 Human raters*

In their study, Johnson et al. (2023) recruited 4 human raters, tasked with assigning ratings to the stories, ranging from 1 (least creative) to 5 (most creative). The instructions for the raters, as taken from the supplementary materials of the original study, can be found in the Appendix I.

*2.2.2 GPT raters*

Following Johnson et al. (2023), we employed 4 artificial raters. To do this, we developed Python code using Google Colab, thus automating the rating process. Using the code, we made four separate calls to the GPT-3.5 API, each one serving as a distinct judge, tasked with rating all of our AI-generated short stories.

To avoid errors, we did not loop through the code four times and instead kept the raters separated in distinct sections. The structure of the code remained the same for each rater to ensure consistency. The code initialises the OpenAI client and then iterates through the stories, sending each one to the API and collecting the returned rating in a list. The list is converted into a dataframe, which is later integrated into a complete dataframe, containing the ratings of all four judges.

We were able to keep the general instructions for the raters concise; however, we needed to put special emphasis on the output format, specifically requesting single-number-only ratings to avoid obtaining in-depth analyses (GPT-3.5 sometimes explained the reasons for a rating even when not asked to). This approach ensured that we obtained the expected output format with high consistency. The full code can be found on OSF (https://osf.io/gcjqv/).

**2.3 Computing Cognitive Networks from Short Stories**

In this work we use textual forma mentis networks (TFMNs) to analyse the structural and conceptual construction of short stories. TFMNs provide a structured representation of mental associations between concepts in a narrative, capturing both structural and emotional links (Stella, 2020). TFMNs are constructed from textual data and break down narratives into syntactic and semantic connections between words to represent how ideas connect within narratives (Improta et al., 2024; Semeraro et al., 2024). TFMNs rely on spaCy, a natural language processing library that analyses text to identify semantic and syntactic relationships between concepts in a given text. SpaCy performs tasks such as sentence splitting, tokenisation, and syntactic parsing, which allow TFMNs to accurately model grammatical and associative relationships within each sentence (Improta et al., 2024). The construction of TFMNs relies on splitting a text into sentences and iteratively performing the following steps on each sentence:

1) **Tokenisation**: Each sentence is split into a set of tokens, i.e. words in our case.

2) **Syntactic parsing:** Using spaCy, each sentence undergoes syntactic parsing to form a syntax tree, identifying connections between words that share syntactic dependencies (e.g. subject and object connected grammatically to the verb) even if they are separated by other elements within the sentence. This provides a considerable advantage over co-occurrence networks (Amancio, 2015), which only create connections between concepts that are located within a specified distance of each other. Consider the following example sentence: "Peter, despite his lactose intolerance and the high cost of milk-based products, loves cheese." By analysing the text on the basis of a syntactic tree, the subject ("Peter") can be linked without problems to the grammatically connected verb and object ("loves cheese") despite the long intermediate phrase ("despite his lactose intolerance and the high cost of milk-based products"). This syntactic dependency would go unnoticed in a simple co-occurrence network (Improta et al., 2024; Semeraro et al., 2024).

3) **Connecting non-stop words on the syntactic tree**: Once the syntactic trees are generated for each sentence, links are established between pairs of non-stopwords (e.g., nouns, verbs, adjectives, adverbs or in general words possessing a meaning of their own). Stop words (such as "and, or this, the") are not considered in this step. Non-stop words are linked if within a distance of 3 nodes on the syntax tree. This approach allows only words that are syntactically close (regardless of how many words are positioned at each syntactic node) to be linked within the TFMN. In the example above, "Peter" is only separated by one intermediate syntactic node from "loves" despite twelve words occupying that syntactic level (Semeraro et al., 2024).

4) **Construction of TFMNs**: For each sentence, TFMNs are built separately, linking syntactically related tokens at a maximum distance of three steps from another (Semeraro et al., 2024). The networks are then merged across sentences to create a unified TFMN for the entire text. This is done by appending all edge lists from each sentence into a unique simple graph, thus creating an edge list representing the whole text (Semeraro et al., 2024).

5) **Normalisation of words**: After constructing an edge list for the entire TFMN, the tokens are normalised for consistency through lemmatisation. Lemmatisation means transferring the specific word into its lemmatised form, which is its base or dictionary form, known as lemma. Thus, different inflections of a word are grouped together to their underlying lemma (e.g. "happiness" and "happier" → lemma "happy"). Tokens that become the same through normalisation are merged together (Semeraro et al., 2024).

6) **Assigning emotional valence**: Using the EmoLex dataset (Mohammad & Turney, 2013), emotional values are assigned to each word, marking them as positive, negative or neutral. In this step, the syntactic parsing supports taking negations into account for assigning emotions. For example, the phrase "not happy" should be evaluated as the opposite of "happy" to correctly grasp the meaning. When emotional scores are computed, by analysing terms based on the syntactic structure, a negation can be properly linked to its associated term, which is then replaced with its antonym as encoded in WordNet, thereby refining the emotional analysis (Improta et al., 2024; Semeraro et al., 2024).

7) **Quantifying emotional features via z-scores**: The emotional features are quantified using z-scores, which measure how rich a given text is in terms of emotional words compared to a null model. To simulate the expected distribution of emotion words, a null model is constructed by randomly sampling words from EmoLex, a psychological lexicon of emotion words commonly used for emotion detection in texts (Semeraro et al., 2024). The z-score for each emotion is calculated by comparing how many words in the text evoke that emotion with how many would be expected if words were chosen randomly from a reference lexicon such as EmoLex. EmoAtlas sets a statistical threshold of |1.96|, meaning that z-scores greater than 1.96 or less than -1.96 indicate that the emotion is significantly more or less frequent than would be expected by random chance, with a significance level of 0.05. By analysing these z-scores, we can obtain a nuanced understanding of the impact of each emotion on narrative creativity (Semeraro et al., 2024).

## 2.4 Extracting Network Features from Short Stories

After constructing a forma mentis network for every story in the dataset, we can extract the following key network features from the syntactic and semantic structure of word associations in texts: Diameter, average shortest path length, clustering coefficient, degree centrality, and PageRank centrality. These measures reveal global aspects of the narrative´s structure, such as its interconnectedness and cohesion (Newman, 2010).

A measure of distance is the **average shortest path length (ASPL)**. It quantifies the efficiency of information flow within a network by calculating the average of the shortest path lengths between all pairs of nodes in a network. A shortest path length is the length of the sequence of links connecting any two nodes with the fewest hops. The average shortest path length refers to the average number of fewest steps needed to link two concepts within a network. A lower ASPL indicates that nodes are more directly reachable from each other, facilitating a faster and more efficient flow of information (Newman, 2010). From a cognitive perspective, ASPL can measure how easy it could be for activation to spread between any two concepts. This was recently investigated in computational studies (Siew, 2019) of the spreading activation model in cognitive psychology (Collins & Loftus, 1975).

The **diameter** of a network represents the length of the longest shortest path between any two nodes. It measures the furthest distance across the network, providing insight into the spread of the network. A larger diameter indicates a more dispersed structure, which can suggest limits to the efficiency of information flow across the network (Newman, 2010). Network diameter was identified as a predictive feature for investigating creativity ratings in a past study (Semeraro et al., 2024).

The **local clustering coefficient** represents the probability that two neighbours of a given node are also neighbours to each other, forming triangles. For the entire network, the mean clustering coefficient is the average of these local coefficients across all nodes. A high clustering coefficient suggests a network with densely connected local structures, which can indicate a modular, clique-based or community-based organisation within the network (Newman, 2010). The local clustering coefficient can be considered as a way for associations between concepts to alter the spreading of activation signals so that two concepts with the same degree (see next paragraph) but different clustering coefficients might undergo different retrieval mechanisms in word confusability tasks (Siew, 2019).

**Degree centrality** can identify the most influential nodes in a network as those with the highest number of connections to other nodes. Degree measures the number of links a given node has with other nodes. Nodes with higher degree centrality are considered to be more central or influential within the network, as they have more direct interactions with other nodes.

This measure helps to identify hubs or key nodes within a network structure (Newman, 2010). Degree centrality can be considered a measure of semantic richness, identifying how many different syntactic and semantic associations a given concept possesses (Semeraro et al., 2024).

Finally, **PageRank centrality** is a measure of a node's importance based on its connections to other influential nodes. Unlike degree centrality, PageRank considers not only the number of incoming links a node has, but also the significance of the nodes linking to it. In terms of probability, PageRank centrality represents the probability that a random walker can visit a given node by either traversing links at random or teleporting across nodes. Nodes with higher PageRank centrality will be visited more often by such a random walker and will, thus, be more influential in the network structure. A node that is connected to highly ranked nodes receives a higher PageRank score, making this measure useful for identifying influential nodes based on their positions within the overall network (Newman, 2010). PageRank centrality has been shown to predict lexical retrieval patterns in experiments with humans (Griffiths et al., 2007).

## 3. Descriptive statistics of the dataset

For our analysis, we employed an interpretable machine learning approach to explore the relationships between network and emotion features and the creativity ratings of short stories. We developed multiple models to analyse the dataset comprehensively as follows:

(i) a model for the human dataset (humans rating human stories);

(ii) a model for GPT-3.5 ratings of human stories;

(iii) a model for the entire GPT-3.5 dataset (GPT-3.5 rating GPT-3.5 stories).

Each model considers only stories rated by all 4 raters and joins them together into one dataset without differentiating between raters. Minimal variation in ratings among the GPT-3.5 raters (Figure 1B and 1C) indicate a high consistency between GPT-3.5 raters.

To enable a clearer and more interpretable analysis of creativity ratings, we decided to transform the original 5 classes of ratings into 3 broader classes. This transformation is justified for several reasons. Firstly, it simplifies the complexity of the data, making patterns and trends easier to identify and interpret. Secondly, it helps mitigate issues related to sparsity within some rating categories, which can enhance the robustness and reliability of our machine learning models. By consolidating the ratings, we reduce noise and potential outliers that may disproportionately influence the results. Thus, we categorised the ratings as follows:

(i) the low class (class 0) includes ratings 1 and 2;

(ii) the middle class (class 1) corresponds to rating 3;

(iii) the high class (class 2) encompasses ratings 4 and 5.

However, as visible in panel C from Figure 1, the dataset of GPT-rating-GPT is completely lacking stories that were rated as 1 or 2, which would have resulted in a 3-class model with an empty class. Thus, we adjusted the categorisation of the ratings for the GPT-rating-GPT dataset in the following way to still allow a 3-class model comparison with the other models:

(i) the low class (class 0) includes rating 3 (since it was the lowest rating assigned);

(ii) the middle class (class 1) corresponds to rating 4;

(iii) the high class (class 2) contains only rating 5.

This transformation provides a more intuitive and cohesive view of the data, facilitating a better understanding of the distinctions between less creative, moderately creative, and highly creative narratives. Figure 2 depicts the distribution of ratings under this new 3-class system, highlighting how this approach achieves a more balanced and comprehensible categorization of creativity ratings.

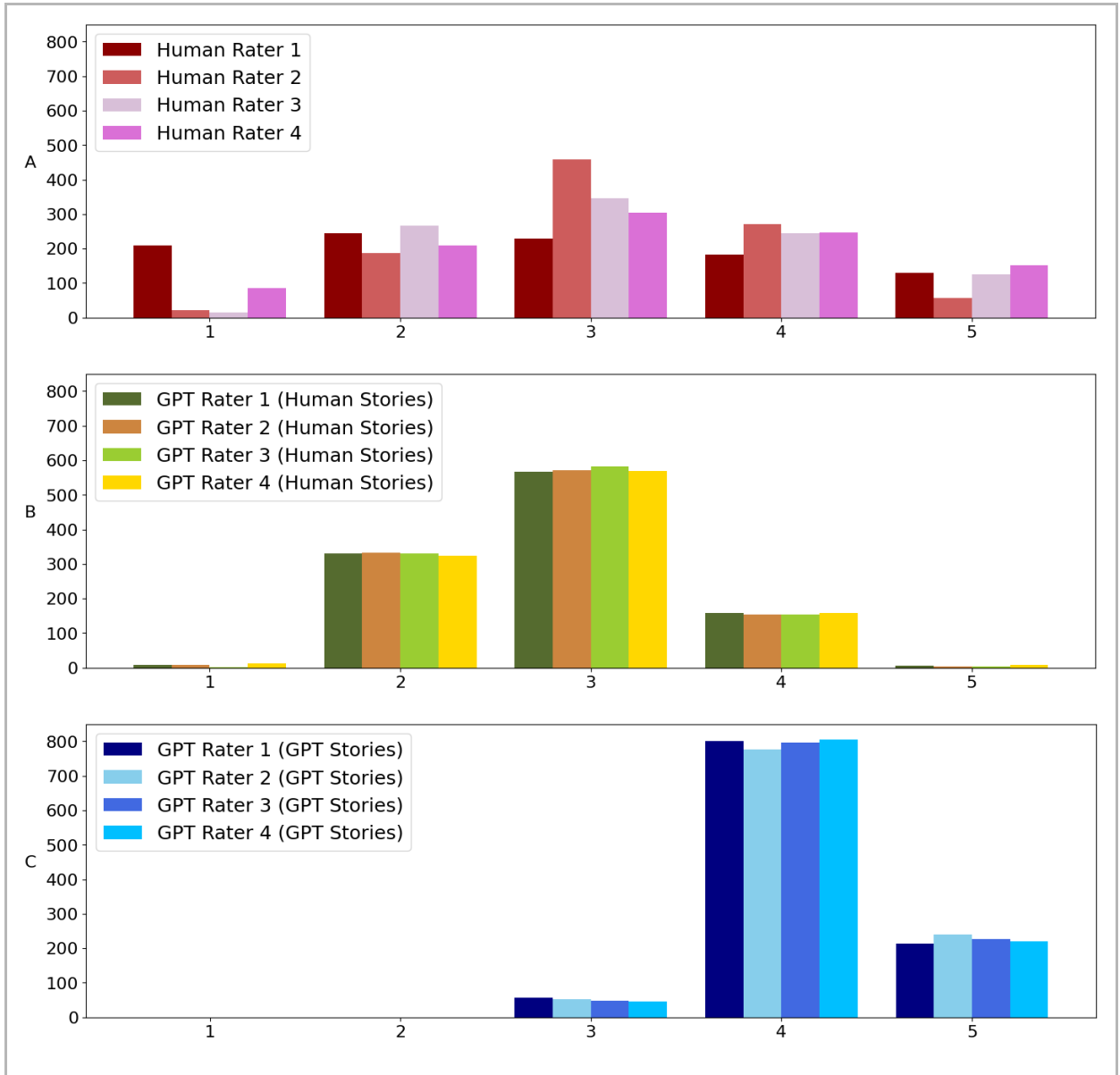

*Figure 1.* Full rating distribution of the three different datasets (1 = very uncreative; 3 = medium creative; 5 = very creative). Panel A displays human ratings for human stories with visible variations across the four human raters. Panel B displays GPT-3.5 rating human stories and panel C shows GPT-3.5 rating GPT-3.5 stories. The GPT-3.5 raters display almost no discernable variation in their rating distribution. Between panel B and C one can notice a clear preference of GPT-3.5 for stories produced by GPT-3.5 themselves rather than human stories, which were rated considerably lower. See a comparison of ratings of individual stories that were rated high in creativity by human raters versus GPT-3.5 in the appendix section IV. The bias of GPT-3.5 in favour of GPT-generated stories is visible from the complete absence of stories rated as 1 or 2 (low creativity) in panel C and the high frequency of rating 4 (high creativity).

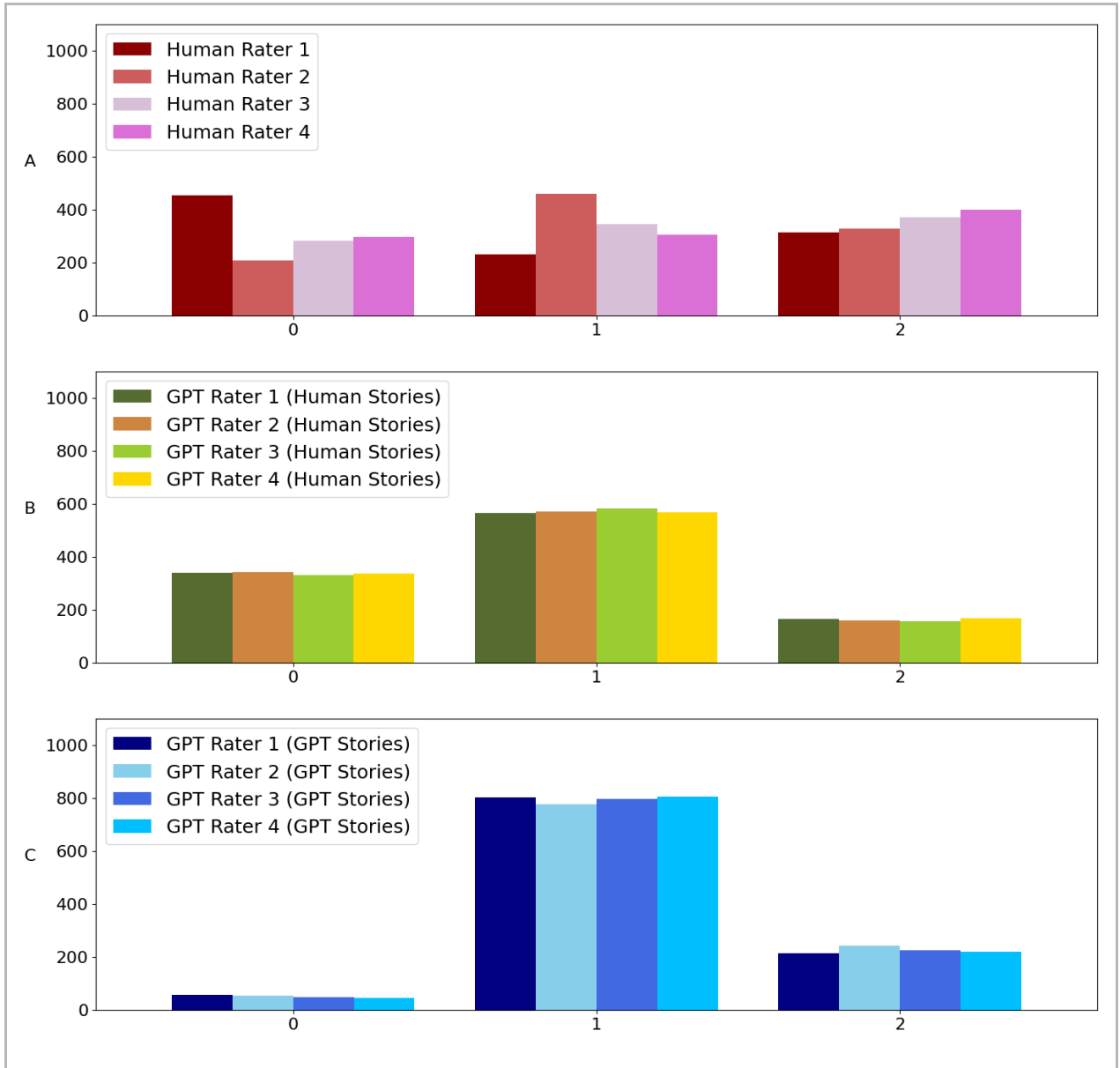

*Figure 2*. Condensed rating distribution of the three different datasets. *0 = low; 1 = middle; 2 = high creativity. For panel A and B (ratings of human stories), the classes are created as follows: class 0 = ratings 1 and 2; class 1 = rating 3; class 2 = ratings 4 and 5. For panel C, the classes were rearranged due to a complete lack of ratings 1 and 2 in the GPT-rating-GPT dataset: class 0 = rating 3; class 1 = rating 4; class 2 = rating 5.*

To test whether the distribution of ratings for human-authored stories differed significantly between human and GPT-3.5 raters, we conducted a Mann-Whitney U test for the reduced dataset, i.e. only stories rated by all four human raters. We found a statistically significant difference between human and GPT-3.5 ratings of human-authored stories (U = 568758.0, $p < 0.001$). Furthermore, we conducted correlation analyses to assess the relationship between the ratings of humans and GPT-3.5 for human-authored stories. Pearson's correlation (r = 0.006, p = 0.848), Spearman's correlation (ρ = 0.009, p = 0.770) and Kendall's Tau correlation (τ = 0.007, p = 0.771) tests all showed no statistically significant relationship between the sets of ratings, meaning that GPT-3.5 ratings did not align with human ratings for the same stories. This indicates that GPT-3.5 and human raters approached evaluating creativity in narratives differently, potentially assigning varying importance to different features of the stories. We will investigate and discuss these potential differences in XAI feature importance analysis via SHAP scores in Section 4.3.

The machine learning model we used is the XGBoost classifier. It operates by employing a gradient boosting framework, where multiple decision trees are built sequentially, and each new tree focuses on correcting the errors made by the previous ones. XGBoost stands out due to its regularisation techniques, which help prevent overfitting, and its efficient handling of missing data, parallel processing, and scalability, making it suitable for large-scale datasets (Chen & Guestrin, 2016). This classifier was selected after a comparative analysis of 12 different classifiers (including logistic regression, decision tree, and bagging), with XGBoost consistently outperforming them in terms of accuracy (see Section 4.2 for the analysis outcome).

To ensure robust model performance, we employed a 4-fold cross-validation. Cross-validation is a robust technique for model evaluation and involves partitioning the dataset into *k* subsets, or folds, where the model is trained on *k-1* folds and tested on the remaining fold. This process is repeated *k* times, ensuring each fold is used as test set once. Cross-validation mitigates overfitting and provides a more accurate assessment of model performance compared to a simple train-test split by utilising the entire dataset for both training and validation (Ferdinandy et al., 2020). By splitting the dataset into four subsets, each classifier was trained on three folds and tested on the remaining fold, rotating this process across all folds. We generated a classification report for the 4-fold cross-validated model, which included error bounds to assess the variability in our predictions. Additionally, we constructed a confusion matrix to visualise the performance of the classifier in distinguishing between different creativity rating levels.

To interpret the results of our models, we incorporated techniques from XAI, focusing on the SHAP (SHapley Additive exPlanations) method to shed light on the inner workings of the XGBoost classifier. SHAP values provide insights into the contribution of each feature to the model's predictions. This helps translate the model's decision-making process into human-interpretable insights. By using SHAP, we gained clarity on how both network and emotion features influence the creativity ratings, identifying the most significant factors that drive these predictions.

This approach not only enhances the transparency of the machine learning model but also aligns with broader XAI objectives of making complex models more interpretable and trustworthy. It is especially important when working with models such as XGBoost, which can behave as a "black box" due to their complexity. Through SHAP, we can understand how various structural and emotional features of the narratives contribute to perceived creativity, deepening our understanding of the underlying mechanisms in creative storytelling (Lundberg & Lee, 2017). This use of XAI ensures that our model's decisions are explainable, fostering greater confidence in its ability to generalise and make accurate predictions in creative contexts.

## 4. Results

In this Section, we present the findings from our study, aimed at predicting creativity levels based on network and emotion features of short stories.

## 4.1 Summary Statistics for Network and Emotion Features

On average, human stories are shorter, with about $\underline{n}_H = 70$ words, while GPT-3.5 stories have an average length of $\underline{n}_{GPT} = 121$ words. Fixing a significance level of 0.05, a Mann-Whitney U test identified a statistically significant difference in story length (U-statistic = 42007.5; $n_H = n_{GPT} = 1071$; $p < 0.001$). We resorted to a non-parametric testing after having qualitatively checked that the data was not distributed according to a Gaussian distribution (e.g., it displayed higher skewness and kurtosis).

By computing a Mann-Whitney U test for the reduced human dataset (only stories rated by all four raters) and GPT3.5-generated stories, we observed significant differences in both network and emotion features between human and GPT3.5-generated stories as shown by p-values well below a threshold of 0.001 (Table 2). Only the features diameter, anger and sadness had a p-value larger than 0.05 which indicates that we cannot assume a statistically significant difference between human and GPT3.5-generated stories for these story features.

| Feature | Statistic | p-value |
| --- | --- | --- |
| ASPL | 588416.0 | < 0.001 *** |
| Clustering_coefficient | 315226.0 | < 0.001 *** |
| Degree_centrality | 571438.5 | < 0.01 ** |
| Diameter | 535019.5 | > 0.05 |
| PageRank_centrality | 475691.0 | < 0.001 *** |
| Anger | 520706.0 | > 0.05 |
| Anticipation | 444272.0 | < 0.001 *** |
| Disgust | 623285.0 | < 0.001 *** |
| Fear | 619380.0 | < 0.001 *** |
| Joy | 301284.5 | < 0.001 *** |
| Sadness | 560438.0 | > 0.05 |
| Surprise | 455986.0 | < 0.001 *** |
| Trust | 376603.5 | < 0.001 *** |

***Table 2***. *Statistics from a Mann-Whitney U test comparing human stories with GPT-3.5 stories.*
*\* <0.05; \*\* <0.01; \*\*\* <0.001*

The Mann-Whitney U test ($s_H = 997$; $s_{GPT} = 1071$) shows that GPT-3.5 stories exhibited significantly higher values in clustering coefficient (U-statistic = 315226.0; $p < 0.001$) and shorter distances (U-statistic = 588416.0; $p < 0.001$) than found in human stories. Furthermore, GPT3.5-generated stories showed higher levels of joy (U-statistic = 301284.5; $p < 0.001$) and trust (U-statistic = 376603.5; $p < 0.001$) while human stories had stronger negative emotions such as fear (U-statistic = 619380.0; $p < 0.001$) and disgust (U-statistic = 623285.0; $p < 0.001$) (Figure 3).

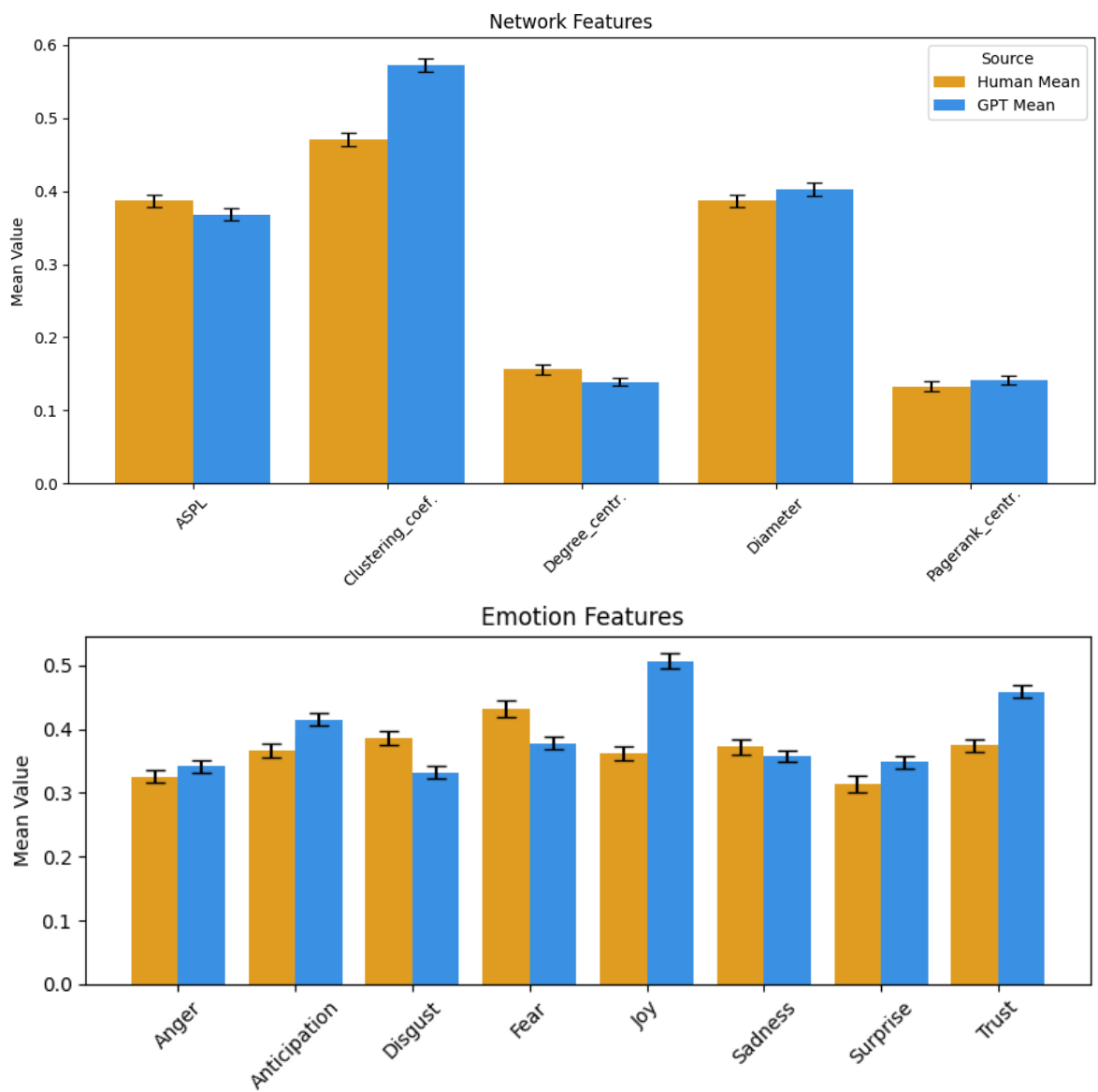

*Figure 3. Comparison of mean values with error bars of network features (top) and emotion features (bottom) for human-generated and GPT-3.5-generated stories.*

## 4.2 Model Performance Evaluation

To evaluate which classifier would be the best for our models, we focused on the accuracy of different classifiers applied to each of the three distinct scenarios: humans rating human-generated stories, GPT-3.5 rating human-generated stories, and GPT-3.5 rating GPT3.5-generated stories. For this reason, we employed a 4-fold cross-validation technique. During this 4-fold cross-validation, the XGBoost classifier emerged as the most effective model due to its superior performance metrics compared to other classifiers (see Table 3). Our evaluation considered twelve different classifiers and their accuracy scores are detailed in Table 3, which lists models from best to worst performance for human ratings. For the human stories, the XGBoost obtained the best accuracy levels (human ratings: 0.617; GPT3.5 ratings: 0.716), followed by the random forest and decision tree classifiers. However, it is noteworthy that for GPT3.5-generated stories, gradient boost (0.757) marginally outperformed XGBoost (0.752). Despite this, to maintain comparability across all models, we consistently used the XGBoost classifier (Chen & Guestrin, 2016; Dietterich, 2000). Notice that we performed this cross-validation on the whole dataset because in this analysis our main aim is not generalisation but rather extracting robust features present in the currently evaluated stories. If we did model selection on a sub-sample of the stories, we would have lost information relative to those specific stories. Hence, to maximise instances of observation of as many stories as possible, we decided to perform model selection on the whole dataset, sacrificing model generalisability but improving the amount of phenomena that we can describe for the current dataset.

| Model | Human-rating-human | GPT-rating-human | GPT-rating-GPT |
|---|---|---|---|
| XG Boost | 0.617 | 0.716 | 0.752 |
| RandomForest | 0.609 | 0.715 | 0.752 |
| Decision Tree | 0.613 | 0.715 | 0.750 |
| Bagging | 0.608 | 0.705 | 0.747 |
| KNN | 0.570 | 0.678 | 0.741 |
| Gradient Boost | 0.581 | 0.645 | 0.757 |
| MLP Classifier | 0.533 | 0.570 | 0.745 |
| Ada Boost | 0.523 | 0.566 | 0.738 |
| Logistic Regression | 0.492 | 0.543 | 0.741 |
| GaussianNB | 0.478 | 0.539 | 0.713 |
| BN | 0.354 | 0.534 | 0.743 |
| SGD Classifier | 0.488 | 0.515 | 0.743 |

*Table 3. Model accuracy of human ratings of human stories (left), GPT-3.5 ratings of human stories (middle) and GPT-3.5 ratings of GPT3.5-generated stories (right) depending on classifier.*

Following the selection of the XGBoost classifier as the most effective model, we employed a 4-fold cross-validation to evaluate each of our three models: human ratings of human-generated stories, GPT-3.5 ratings of human-generated stories, and GPT-3.5 ratings of GPT3.5-generated stories. The classification reports, summarised in Tables 4, 5, and 6, provide insights into the precision, recall, F1 scores, and accuracy, including error bounds for each class. We also tried out the random forest classifier to compare its performance with XGBoost. The results were highly similar across both classifiers, with only minor differences in accuracy. This consistency suggests that the predictive features of our models are robust, regardless of the specific machine learning technique applied.

For the human-rating-human model (Table 4), the classification report indicated for class 2 (high creativity) an accuracy of $0.62 \pm 0.02$, precision of $0.69 \pm 0.04$ and recall of $0.71 \pm 0.04$. The precision and recall for class 0 and 2 were higher than those for class 1 (precision $0.50 \pm 0.02$; recall $0.48 \pm 0.03$), indicating the model's better performance in identifying less and more creative stories in comparison to mid creativity. The ROC AUC score of $0.78 \pm 0.02$ suggests a strong ability of the model to discriminate between different levels of creativity. The GPT-rating-human model (Table 5), on the other hand, demonstrated a weighted average precision, recall, and F1 score of 0.71, with an accuracy of $0.72 \pm 0.01$, precision of $0.60 \pm 0.01$, and recall of $0.55 \pm 0.03$ for class 2. The ROC AUC score of $0.82 \pm 0.01$ indicates excellent discriminatory power. For the GPT-rating-GPT model (Table 6), the classification report revealed a weighted average precision, recall, and F1 score of 0.73, with an accuracy of $0.75 \pm 0.01$. However, the model struggled with mid creativity ratings (class 0 in this case, consisting of rating value 3), reflecting its low frequency in the dataset. This class achieved a precision of $0.30 \pm 0.08$ and a recall of $0.10 \pm 0.02$, significantly lower than the other classes. Despite this, the model's overall performance, with a ROC AUC score of $0.68 \pm 0.01$, was relatively robust.

| Class | Precision | Recall | f1-score |
|---|---|---|---|
| 0 | 0.67 ± 0.06 | 0.66 ± 0.06 | 0.66 ± 0.02 |
| 1 | 0.50 ± 0.02 | 0.48 ± 0.03 | 0.49 ± 0.03 |
| 2 | 0.69 ± 0.04 | 0.71 ± 0.04 | 0.70 ± 0.01 |
| | | | |
| Accuracy | 0.62 ± 0.02 | | |
| roc_auc | 0.78 ± 0.02 | | |
| Macro avg | 0.62 | 0.62 | 0.61 |
| Weighted avg | 0.62 | 0.62 | 0.62 |

*Table 4. Cross validated (4-fold) classification report with error bounds for the human ratings of human-generated stories. Class 0 = low creativity; class 1 = mid creativity; class 2 = high creativity.*

| Class | Precision | Recall | f1-score |
|---|---|---|---|
| 0 | 0.75 ± 0.01 | 0.72 ± 0.01 | 0.73 ± 0.01 |
| 1 | 0.73 ± 0.01 | 0.76 ± 0.01 | 0.74 ± 0.01 |
| 2 | 0.60 ± 0.01 | 0.55 ± 0.03 | 0.58 ± 0.02 |
| | | | |
| Accuracy | 0.72 ± 0.01 | | |
| roc_auc | 0.82 ± 0.01 | | |
| Macro avg | 0.69 | 0.68 | 0.68 |
| Weighted avg | 0.71 | 0.72 | 0.71 |

*Table 5. Cross validated (4-fold) classification report with error bounds for the GPT-3.5 ratings of human-generated stories. Class 0 = low creativity; class 1 = mid creativity; class 2 = high creativity.*

| Class | Precision | Recall | f1-score |
|---|---|---|---|
| 0 | 0.30 ± 0.08 | 0.10 ± 0.02 | 0.15 ± 0.03 |
| 1 | 0.81 ± 0.01 | 0.88 ± 0.01 | 0.84 ± 0.01 |
| 2 | 0.54 ± 0.02 | 0.45 ± 0.02 | 0.49 ± 0.01 |
| | | | |
| Accuracy | 0.75 ± 0.01 | | |
| roc_auc | 0.68 ± 0.01 | | |
| Macro avg | 0.55 | 0.48 | 0.50 |
| Weighted avg | 0.73 | 0.75 | 0.74 |

*Table 6. Cross validated (4-fold) classification report with error bounds for the GPT-3.5 ratings of GPT3.5-generated stories. Class 0 = low creativity; class 1 = mid creativity; class 2 = high creativity.*

The confusion matrices (Figure 4) for each model further illustrate their ability to correctly classify different creativity levels. For the human-rating-human model, the confusion matrix performs well in identifying "high" creativity with 1006 correct predictions. In contrast, the GPT-rating-human model performs better for the "mid" class (1744 correct predictions) which was also the most frequent class in this model. The GPT-rating-GPT model also performs best for the "mid" class (2795 correct predictions) but struggled with predicting "low" creativity accurately, with only 21 correct predictions and many misclassifications into "mid" classes. This likely results from the smaller sample size of low ratings.

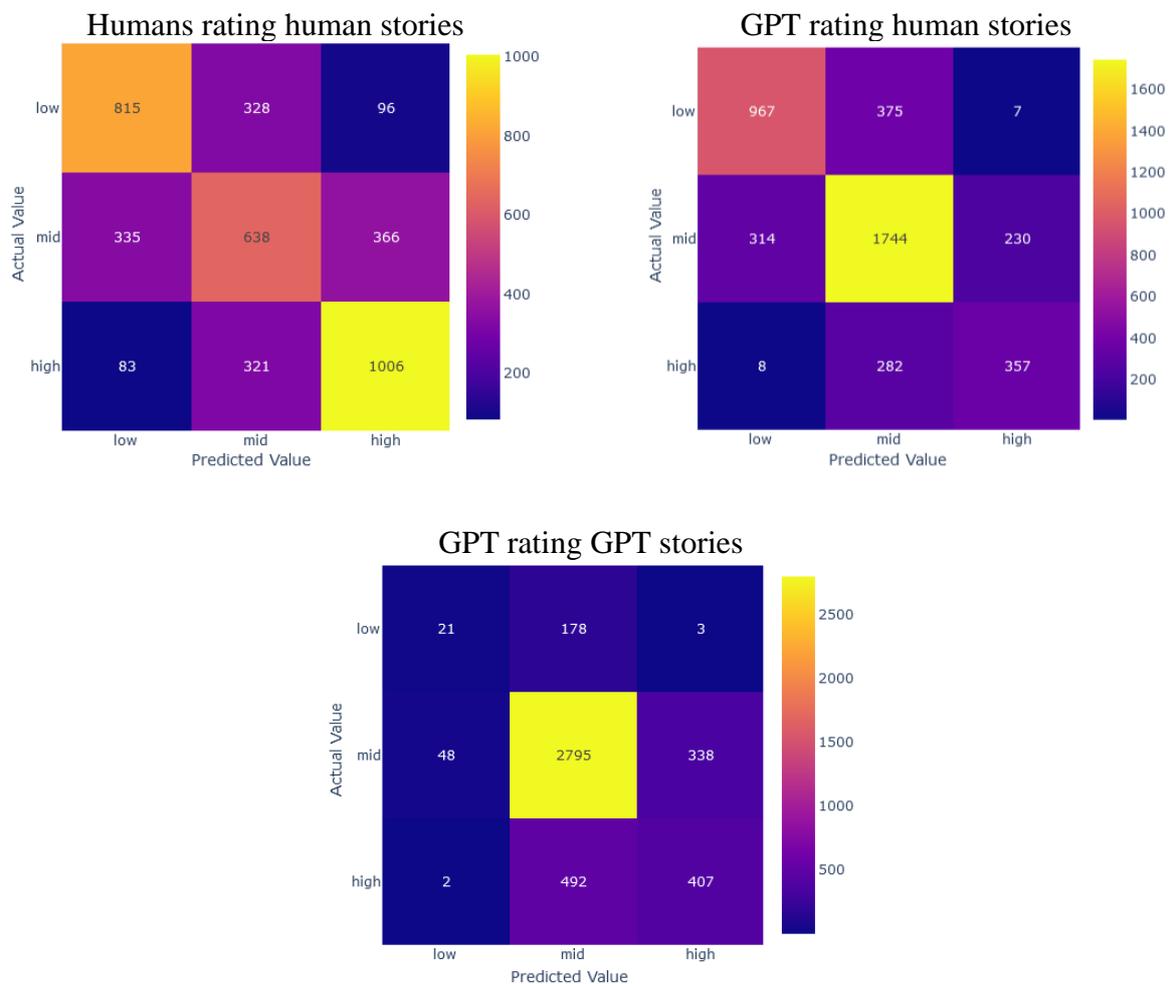

*Figure 4.* 4-fold cross validated confusion matrices for the three different models. For the human-generated stories: low creativity = rating 1 or 2; mid creativity = rating 3; high creativity = rating 4 or 5. For the GPT-rating-GPT model: low creativity = rating 3; mid creativity = rating 4; high creativity = rating 5.

**4.3 Feature Importance Analysis**

In our exploration of feature importance for predicting creativity ratings, we employed SHAP value plots (Lundberg & Lee, 2017) using the SHAP package in Python (Lundberg et al., 2019) to visualise the influence of various network and emotion features across three rating models: the human-rating-human model, the GPT-rating-human model, and the GPT-rating-GPT model. The SHAP plots display the average impact on model output magnitude for each feature within these models. In Figure 5, panel A shows the human model, where network features like PageRank centrality and degree centrality are most influential, particularly for class 2 (high creativity). Emotion features such as disgust and trust also play significant roles, indicating that certain emotional tones correlate with higher creativity ratings. Figure 5, panel B depicts the GPT-rating-human, which balances both network and emotion features. PageRank centrality and degree centrality remain crucial, but average shortest path length (ASPL), sadness, and clustering coefficient also have significant impacts, reflecting a nuanced consideration of structural and emotional aspects in creativity prediction. Figure 5, panel C presents the GPT-rating-GPT model, which heavily relies on emotion features like anger, anticipation, and joy, especially for class 1 (middle creativity) and class 2 (high creativity). This highlights GPT-3.5´s preference for emotional content over network features. The GPT-rating-GPT model's distinct feature importance profile underscores its differing approach in creativity assessment compared to human raters. These results suggest that the human-rating-human model prioritises structural features, while the GPT-rating-GPT model focuses more on emotional cues. This emphasis on emotional cues in GPT-3.5´s stories is further supported by a qualitative view on the narratives. A closer reading reveals a reliance on vivid, pictorial language rich in positive emotional content. This perception is aligned with the findings from Section 4.1, where GPT-3.5 stories demonstrated statistically significant higher z-scores for emotions such as joy, anticipation, and trust compared to human-authored stories. This distinction underscores GPT-3.5´s tendency to enhance its narratives with pronounced emotional tones, which may contribute to its differing approach to creativity assessment, favouring emotional features over structural ones.

In contrast, the GPT-rating-human model assigns considerable weight not only to emotional features but also to network features, thus aligning more closely with the human raters than the GPT-rating-GPT model. However, the difference found between human and GPT-3.5 ratings of human-authored stories (see Section 3) suggests that GPT-3.5´s evaluations are still distinct from those of human raters, even when rating human-authored stories. What about the evaluation criteria behind these ratings? Interestingly, despite scores attributed to stories differing between GPT-3.5 and humans, SHAP plots (Fig. 5 and Fig. 6) highlight that GPT-3.5 appears to rely on features more similar to human criteria when assessing human-authored stories than when assessing GPT-generated stories.

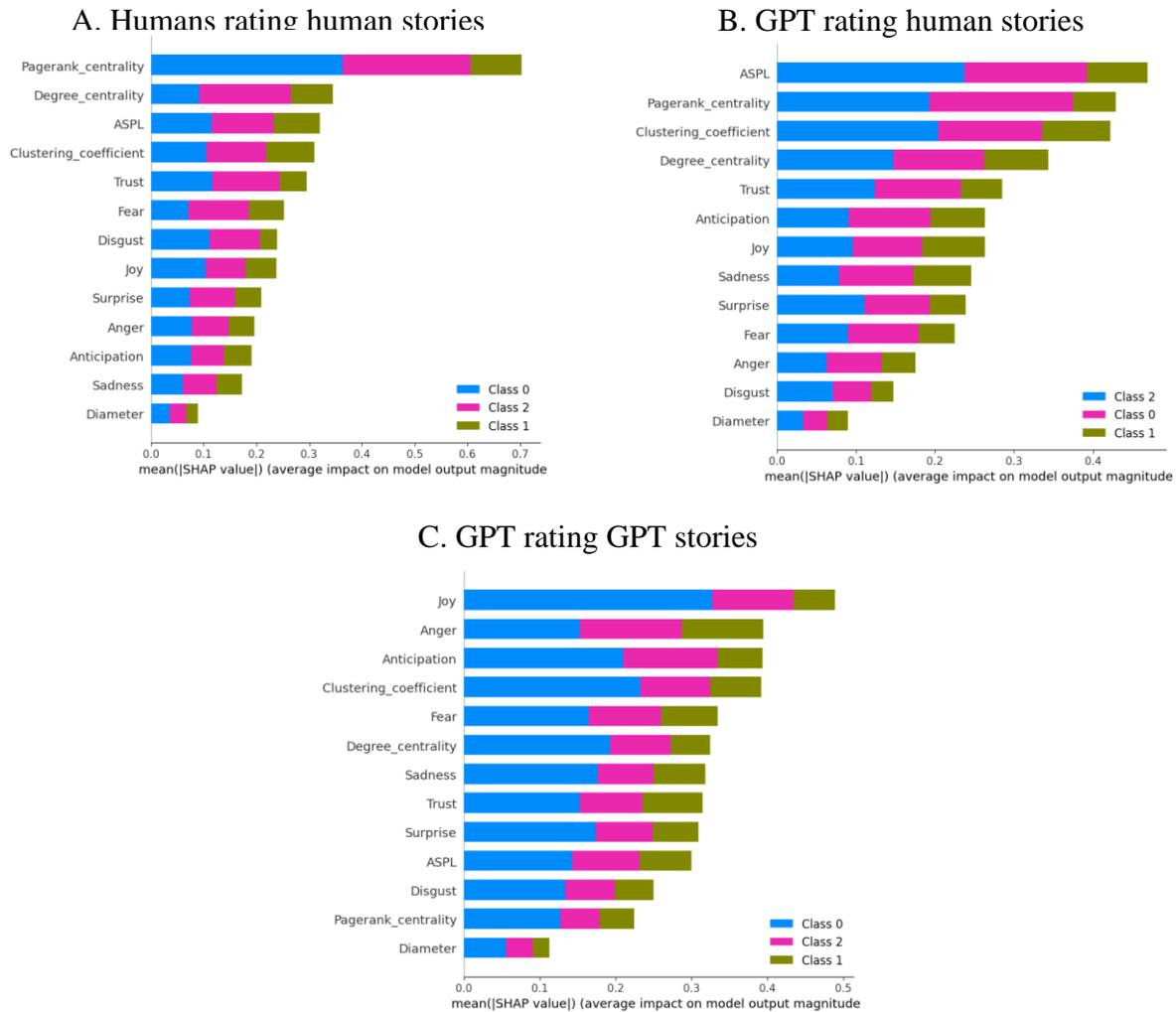

*Figure 5. Feature importance plots for the three different models. For the human-generated stories: class 0 = low creativity (ratings 1 or 2); class 1 = mid creativity (rating 3); class 2 = high creativity (ratings 4 or 5). For the GPT-rating-GPT model: class 0 = low creativity (rating 3); class 1 = mid creativity (rating 4); class 2 = high creativity (rating 5).*

The panels in Figure 6 depict the feature importance of the three models as beeswarm plots. The human-rating-human model (Fig. 6, left column) and GPT-rating-human (Fig. 6, middle column) depict a similar pattern as seen in Figure 5 with PageRank centrality, ASPL, and degree centrality remaining prominent in feature importance across all classes. In contrast, the GPT-rating-GPT model (Fig. 6, right column) shows a distinct approach, emphasising emotion features such as anger, anticipation, fear, and joy.

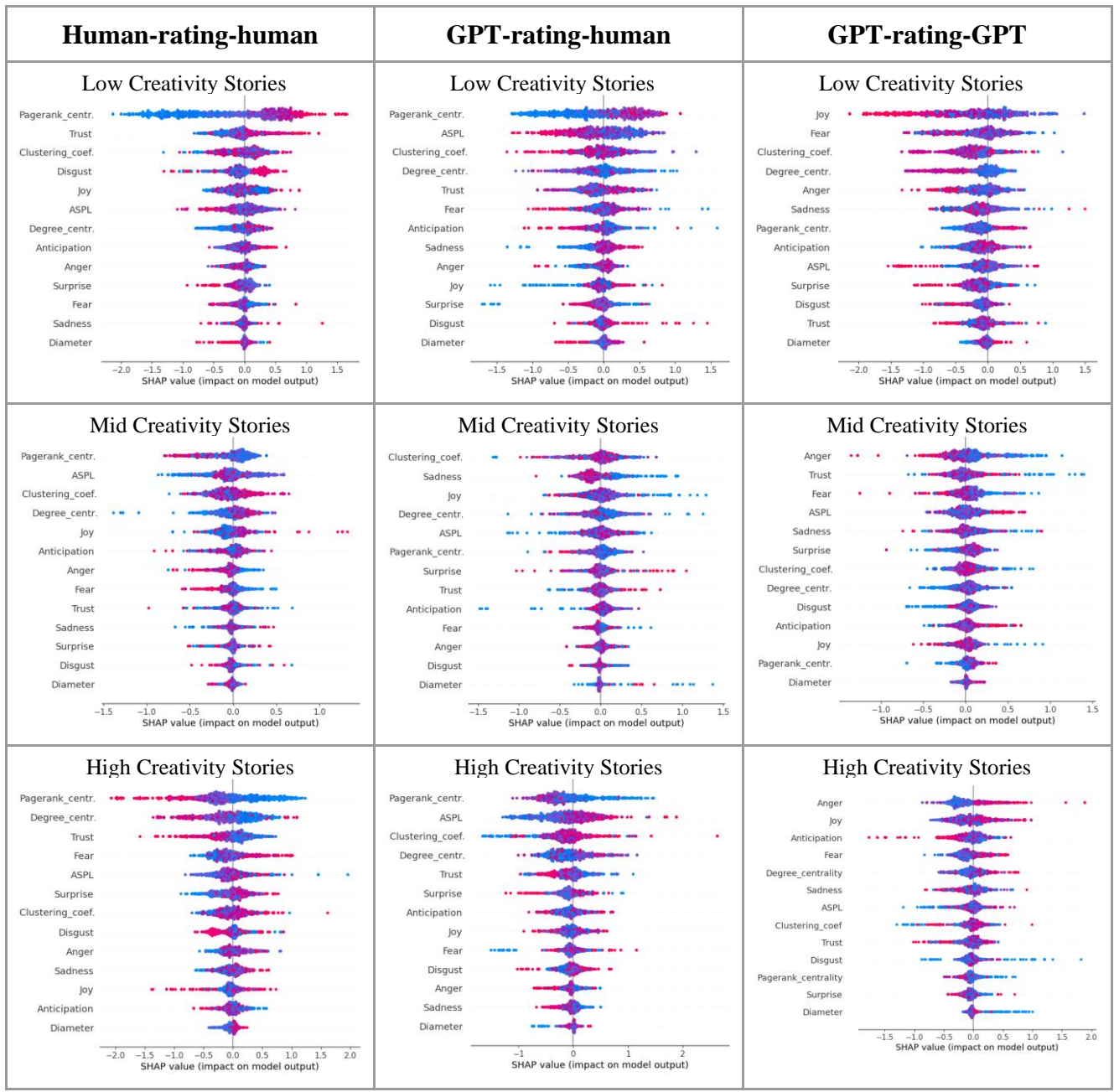

*Figure 6.* Feature importance beeswarm plots for the human-rating-human model (left), the GPT-rating-human model (middle) and the GPT-rating-GPT model (right) separated by class (low, mid and high creativity stories). Feature value intensity is indicated as weak (blue), moderate (purple) and strong (pink) intensity.

The difference in feature importance assigned by GPT-3.5 raters for either human-generated stories (Fig. 6, left) or GPT3.5-generated stories (Fig. 6, right) is striking given that GPT-3.5 was unaware of the origin of the stories. We expected the GPT-3.5 raters to use the same features or characteristics for assigning creativity scores to all stories equally without an explicit bias for stories generated by GPT-3.5. However, our findings indicate that the GPT-3.5 raters used different features with varying degrees of importance for assigning creativity ratings to either the human or GPT-3.5 stories. Since there was minimal variation in ratings across the four GPT-3.5 raters within each model, one can assume that these differences are not merely due to random rating assignments.

When comparing the performance of GPT-3.5 for rating human-generated stories in contrast to GPT-3.5-generated stories, an intriguing pattern emerges. When rating human-generated stories, GPT-3.5 demonstrates the ability to assess human creativity in a manner more similar to human raters than when assessing GPT-generated creativity (Fig. 6, left and middle columns). In this context, GPT-3.5 relies on a combination of network and emotion features with network features (e.g., PageRank centrality, clustering coefficient, and degree centrality) showing stronger feature importance than emotion features. However, despite these seeming similarities on assessment criteria, the correlation coefficients reported in Section 3 indicate no statistically significant alignment between GPT-3.5 and human ratings. This finding suggests a notable gap between GPT-3.5´s outputs for creativity assessment and those of human raters. Thus, users need to be cautious about using GPT-3.5 for rating the creativity of a story since GPT-3.5´s ratings might diverge considerably from human ratings.

When comparing the GPT-rating-human model to GPT-3.5 generating and evaluating its own stories, a striking difference arises between the two sets of GPT raters. The model for GPT-rating-GPT stories relies more heavily on emotion features such as joy, anger, and anticipation, while structural features like PageRank centrality become less relevant compared to how both human- and GPT-raters assess human stories. This divergence implies that GPT-3.5 operates under a different internal framework when generating creative content compared to when it is rating human creativity. Such a shift calls for caution when working with AI to create content. While transformers like GPT-3.5 can be reliable in assessing the creativity of human stories, their process for generating creative stories appears to differ from the human creative process. The very features that GPT-3.5 emphasises when creating stories are not the same features that human raters, or even GPT-3.5 itself when rating human output, consider important. This discrepancy may lead to GPT-3.5 producing stories that lack the structural complexity typically associated with human creativity (Green et al., 2024).

## 5. Discussion

### 5.1 Interpretation of Findings

Our study reveals significant insights into the structural and emotional differences between human-authored and GPT3.5-generated stories, with implications for how these narratives are perceived and rated for creativity. The analysis showed that GPT-3.5 stories tend to be longer (average of 121 words) compared to human stories (average of 70 words). This length difference is accompanied by higher clustering coefficients in GPT3.5-generated stories, indicating more interconnected narrative structures. On an emotional level, GPT3.5-generated stories exhibited higher levels of positive emotions such as joy and trust, whereas human stories contained more negative emotions like fear and disgust. These distinctions were statistically significant, with p-values well below the 0.05 threshold.

The SHAP value plots provided insights into the influence of various features across the models. For the human model, network features like average PageRank centrality and degree centrality were most influential, particularly for high creativity ratings. Higher average values of PageRank centrality correlate with stories that are more interconnected, looping back and linking concepts to one another. This type of structure is syntactically more interconnected and it places concepts at a lower network distance, potentially corresponding to more rigid and highly connected network structures, associated in past works to lower levels of creativity (Kenett, 2019). Conversely, stories with lower average PageRank centrality venture further without linking back as much, making them syntactically less cohesive and with longer sequences of syntactic relationships, linking concepts remaining at a longer network distance. This structure correlates with higher creativity, aligning with the idea that novelty and unpredictability are key components of creative expression (Griffiths et al., 2007; Zhang et al., 2022).

Emotional features of stories, such as disgust and trust, also played significant roles, suggesting that certain emotional tones correlate with higher creativity ratings. Prior studies underscore the complex relationship between emotional expression or moods and creativity levels of narratives. For instance, a study on the EmoAtlas package also used in the present research (Semeraro et al., 2024) demonstrates that expressing emotions within texts can influence creativity assessments, especially when evoking distinct emotional tones. By mapping emotions through z-scores and emotional-syntactic networks, EmoAtlas highlights how creative expression can be channelled through specific emotional associations (Semeraro et al., 2024). This aligns with our findings, where emotions like trust and surprise increased creativity levels. Further support for this comes from an exploration of computational linguistic indicators of creativity in writing by Zedelius and colleagues (2018). This study found that emotionally expressive language, especially when rich in distinctive tone and voice, is predictive of higher creativity ratings (Zedelius et al., 2018).

In addition, studies on mood and emotion regulation have provided insights into how the moods of individuals during the creative process can impact the output and creativity level of the narrative. Activating positive emotions, such as happiness, during the writing process have been found to enhance the creativity of the text, whereas neutral moods or deactivating positive emotions did not. Deactivating negative emotions like anxiety or fear tend to correlate with lower creative output (Baas et al., 2008). Lastly, work on emotion regulation and creativity in storytelling suggests that effective emotional regulation correlates with higher originality in narratives. This was particularly the case when maladaptive emotional regulation strategies were minimised. This suggests that effective, adaptive regulation (such as acceptance) might support higher levels of originality in narratives, likely by fostering a balanced emotional state beneficial for creative thinking (Kopcsó & Láng, 2017).

In the study at hand, aside from emotional features we found a significant impact of network features on the perceived creativity levels of short stories. Network features were especially relevant for predicting creativity scores of the human-rating-human model. In contrast, the GPT-rating-GPT model relied heavily on emotion features like anger, anticipation, and joy. This preference for emotional content over network features underscores GPT-3.5´s distinct approach to creativity assessment compared to human raters. This divergence from human ratings, which focused more on structural features, underscores the different evaluative frameworks used by human and AI raters.

### 5.2 Contributions to Knowledge - AI and Creativity

This study advances our understanding of creativity assessment in significant ways. By focusing on the prediction of creativity levels in short stories through network and emotion features, our research contributes to the broader discourse on AI's role in creativity assessment. It also highlights important distinctions between how GPT-3.5 perceives and generates creative content compared to human individuals.

In the literature, there is a growing integration of transformer models in creativity assessments. Sun and colleagues (2024) explore the use of semantic distance and machine learning to predict human ratings of creativity in the Alternative Uses Task (AUT). Their study found that contextual semantic models, such as GPT-3 and RoBERTa, were better aligned with human ratings than non-contextual models like GloVe. These results highlight the promising new avenue for using transformers in capturing complex, context-dependent aspects of creativity (Sun et al., 2024).

Similarly, Patterson and colleagues (2023) developed the AuDrA platform, which uses deep learning techniques to assess the visual creativity of drawings. Their model was trained on a large dataset of sketches and demonstrated good correlation with human creativity ratings of the same drawings, outperforming simple metrics for creativity rating such as elaboration. This work emphasises the possible applications of machine learning in automatically evaluating creativity during visual tasks or in visual products (Patterson et al., 2023).

A key finding from this study is that GPT-3.5 ratings for human-authored stories diverged significantly from human ratings. This is indicated by a statistically significant difference in the rating distributions between GPT-3.5 and human raters (see Section 3) and a lack of significant correlations between the two (see Section 3). This highlights that GPT-3.5´s assessments of the human-authored stories did not align with human judgements of creativity. The SHAP plots (Fig. 5 and Fig. 6) also indicate that structural network features of syntactic/semantic relationships between words in stories are relevant for investigating creativity assessments in humans and GPT-3.5: PageRank centrality, clustering coefficient and degree centrality were just as crucial for high creativity ratings in human-generated stories rated by GPT-3.5 as when rated by humans. This finding underscores the significant limitation in using GPT-3.5 as an evaluation tool for human creativity. While GPT-3.5 may capture certain structural and emotional patterns that humans intuitively consider, it does not fully align with human evaluative criteria for creativity. Therefore, the present study emphasises the need for caution in employing GPT-3.5 for creativity assessment in educational or professional settings, where ratings by GPT-3.5 may not reflect human judgements accurately.

Furthermore, when exploring how GPT-3.5 evaluates its own generated content in contrast to human-authored stories, a markedly different picture emerges. While structural features remain important for evaluating human creativity, the GPT-rating-GPT model assigns far less weight to these features when assessing its own stories. Instead, it shifts focus on emotional features such as joy, anger and fear (Figure 6, right column). This suggests that GPT-3.5´s concept of "creativity", particularly during the process of creative generating, is more closely related to eliciting emotional responses rather than maintaining structural novelty or coherence. This impression is strengthened when looking at the individual stories generated by GPT-3.5 that were rated high in creativity by GPT-3.5 itself. These stories tend to be picturesque and focus on emotion-heavy adjectives, as seen from a comparison of the ten most creatively rated stories by human participants and GPT-3.5, found in appendix IV.

Current research about the level of emotion AI models such as GPT-3.5 and GPT-4 can demonstrate suggests that these models can approximate human emotional responses to some extent, but its understanding remains surface-level, particularly in lower-level models like GPT-3.5 (Patel & Fan, 2023; Tak & Gratch, 2023). More recent models like GPT-4 show advancements in recognising and labelling emotions. However, they still tend to fall short of the nuanced empathy inherent in humans (Patel & Fan, 2023; Tak & Gratch, 2023).

The shift in feature importance from network features to a higher importance of emotion features suggests that GPT-3.5 operates under a different cognitive framework when generating content versus when evaluating content. This observation raises important questions about the ability of AI models to generate content that aligns with human notions of creativity. Green et al. (2024) argue that AI such as GPT-3.5 lacks the internally directed attention and goal-oriented processes that are fundamental to human creativity. This concern becomes particularly pronounced when AI models are used to generate creative content without human intervention, as the resulting narratives tend to be less coherent in structure, focusing on emotional engagement at the expense of narrative coherence. Thus, while GPT-3.5 may provide some insights into creativity evaluation, its application both in assessing creativity and generating creative content should be approached with caution. GPT-3.5 does not yet capture the nuanced structural and conceptual complexity that characterises human creativity.

**5.3 Limitations**

At the current time of writing up this manuscript, GPT-3.5 is not available any more for use via the web interface but it remains available in the GPT store and for usage via API, thus still enabling scientific experiments.

While our study provides valuable insights, several limitations should be acknowledged. Firstly, as GPT-3.5 was consistently rating its own stories high in creativity, this resulted in a very small sample size of low-rated GPT3.5-generated stories. This resulted in a more biassed performance of the XGBoost model with many misclassifications in the confusion matrix. Thus, a more balanced dataset in terms of rating frequency would be necessary to improve model reliability.

Additionally, GPT-3.5´s behaviour may differ from more advanced models such as GPT-4, which has been shown to exhibit improvements in nuanced reasoning and evaluation tasks (Espejel et al., 2023). This suggests that our findings may be specific to GPT-3.5 and may not reflect the behaviour of more recent models.

Furthermore, we condensed the dataset to consist of only stories that were rated by all four raters and calculated the model across all raters, disregarding variations between the four distinct human raters. Alternative statistical approaches, such as linear mixed models, could be used to account for the different behaviours of individual raters to minimise potential biases introduced by rater-specific tendencies. Linear mixed models allow for the inclusion of random effects for individual raters, enabling control over rater variability and offering a more accurate analysis of rating patterns (Hox et al., 2017).

**5.4 Future Directions**

Future research could delve deeper into the potential biases present in AI evaluations of creative content. Our findings indicate a clear bias in GPT-3.5´s ratings, favouring stories written by GPT-3.5 over those authored by humans. Addressing this bias is crucial for developing more accurate AI evaluative systems.

One intriguing future direction involves instructing GPT-3.5 to deliberately write uncreative stories. By having these intentionally uncreative stories rated by both GPT-3.5 and human raters, researchers can investigate whether GPT-3.5 can deliberately produce and accurately recognize lower creativity levels. This approach would provide insights into GPT-3.5´s ability to follow specific creative directives and its consistency in evaluating creativity across different narrative qualities.

Another promising area for future research is to test the models and findings across different languages. Examining whether the same structural and emotional features are relevant for stories written in languages other than English would help determine if these features are universally influential in creativity assessments. This cross-linguistic investigation would contribute to a more comprehensive understanding of creativity and the development of AI models that are effective across diverse linguistic contexts.

**Conclusion**

The findings of this study quantify different patterns characterising creativity assessment of short stories in humans and simulated raters (GPT-3.5). By considering textual stories as structured data, i.e. textual forma mentis networks of syntactic/semantic/emotional relationships between words, this study uses explainable AI to identify which features lead to the prediction of human and GPT-3.5´s ratings of short stories. Our findings demonstrate that in producing creativity ratings of human stories, human raters place emphasis on structural semantic network features (e.g. degree centrality and PageRank centrality) in conjunction with using the emotional content of the story (e.g., its elicited joy or trust). Importantly, similar patterns are found also in GPT-3.5. However, when GPT-3.5 rates its own GPT-based stories, the LLM shows a marked reliance on emotional features, suggesting an inclination towards affective attributes over structural ones in the self-assessment of creativity.

This discrepancy points to limitations in current LLM models' interpretative frameworks for creativity, as these models exhibit an emotion-centric approach rather than assessment heuristic based on semantic/syntactic association structure. Such findings underscore the importance of advancing LLMs to enhance their alignment with human evaluative criteria when called to produce "creative" content and rate it.

**Synthetic dataset of creative stories**

The synthetic dataset of 1071 stories generated by GPT-3.5 with their corresponding ratings, as used in this study, can be found publicly available on OSF. This dataset is available for future research and can be accessed via the following link: https://osf.io/gcjqv/

# Appendix

## I. Instructions for Raters

In this task, participants were shown 3 words ("stamp", "letter", "send"), and were asked to write a short story that included all 3 words. They were told to be imaginative and creative when writing their stories.

You will first be shown just 10 stories, and you will have to first rank these 10 stories from least creative to most creative, and then give each of them a rating from 1 (very uncreative) to 5 (very creative). After this you will be shown the rest of the stories and you will have to give each of these a rating for creativity too.

When rating the stories, try not to focus too much on the length of the story, or how good the English is, but consider the overall creativity of the story. You may wish to consider how creatively the 3 words were used, how emotive, descriptive, or humorous the story was, and how much it "came alive".

> Scale:
> 1: Very Uncreative
> 2: Uncreative
> 3: Undecided
> 4: Creative
> 5: Very Creative.

(*Johnson et al., 2023, Supplementary Materials*)

**II. Comparative Analysis for Human-generated and GPT-generated Stories**

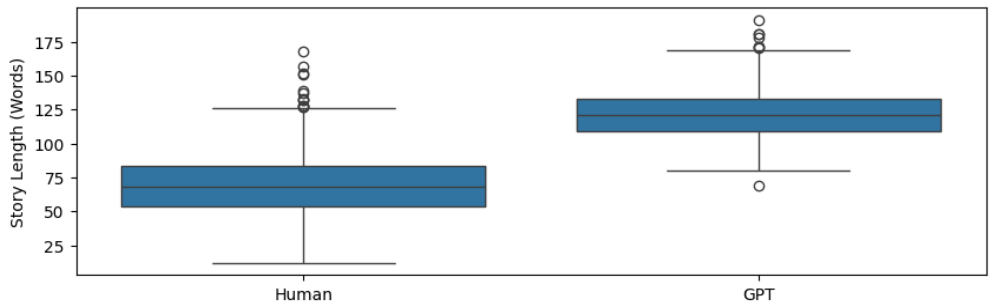

*Figure 7*. Boxplot comparing story length measured in number of words between human-authored and GPT3.5-generated stories with human stories generally shorter than those generated by GPT-3.5.

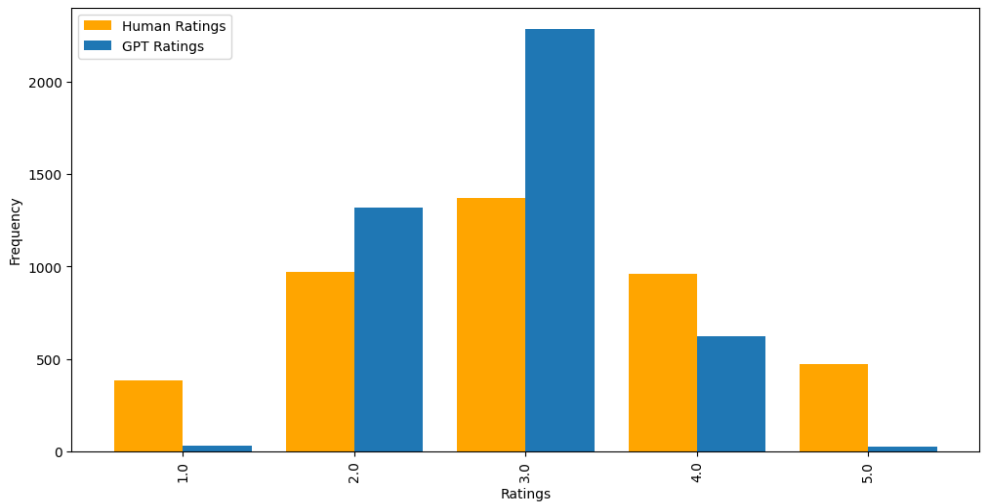

*Figure 8*. Comparative distribution of ratings for human-authored stories by human raters (orange) and GPT-3.5 raters (blue). Notice a clear trend of GPT-3.5 for more medium ratings (2-4) and very few extreme ratings (1 or 5).

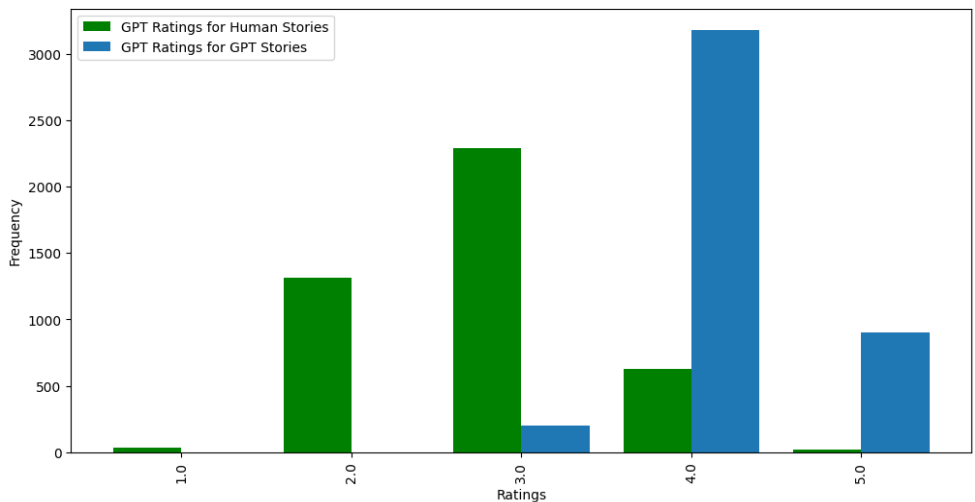

*Figure 9*. Comparative distribution of ratings by GPT-3.5 raters for human-authored (green) and GPT-generated stories (blue). Notice a clear preference of GPT-3.5 in favour of GPT-generated stories with strikingly higher ratings (4 and 5) and complete lack of ratings 1 and 2. In contrast, human-authored stories were rated as less or medium creative (frequent ratings of 2 and 3).

## III. Summary Statistics of Network and Emotion Features of the Dataset

| **Network features in human-generated stories** | | | | | |
|---|---|---|---|---|---|
| | ASPL | Clustering | Degree_cent | Diameter | PageRank_cent |
| mean | 0.386972 | 0.470768 | 0.156140 | 0.386760 | 0.133139 |
| std | 0.131258 | 0.137863 | 0.111265 | 0.135049 | 0.106944 |
| 25% | 0.301478 | 0.391096 | 0.085784 | 0.300000 | 0.066092 |
| 50% | 0.371915 | 0.479688 | 0.128284 | 0.400000 | 0.105072 |
| 75% | 0.463094 | 0.562771 | 0.188933 | 0.500000 | 0.157407 |
| **Network features in GPT3.5-generated stories** | | | | | |
| | ASPL | Clustering | Degree_cent | Diameter | PageRank_cent |
| mean | 0.368177 | 0.572487 | 0.139017 | 0.402088 | 0.141838 |
| std | 0.130123 | 0.137215 | 0.088125 | 0.149182 | 0.099014 |
| 25% | 0.276920 | 0.489863 | 0.083592 | 0.272727 | 0.083131 |
| 50% | 0.347736 | 0.578193 | 0.121928 | 0.363636 | 0.118250 |
| 75% | 0.441351 | 0.661575 | 0.171247 | 0.454545 | 0.171329 |

*Table 7*. Summary statistics of the scaled network features of the human dataset (above) and the GPT-3.5 dataset (below). Minimum and maximum values are omitted because all features were scaled between 0 and 1, resulting in constant values of 0 for minimum and 1 for maximum values.

| **Emotion features in human-generated stories** | | | | | | | | |
|---|---|---|---|---|---|---|---|---|
| | Anger | Anticipation | Disgust | Fear | Joy | Sadness | Surprise | Trust |
| mean | 0.325442 | 0.366624 | 0.385550 | 0.432085 | 0.362140 | 0.372512 | 0.314179 | 0.374122 |
| std | 0.152243 | 0.174006 | 0.173384 | 0.203495 | 0.181201 | 0.193006 | 0.192256 | 0.171248 |
| 25% | 0.200671 | 0.202309 | 0.241783 | 0.280499 | 0.218370 | 0.244864 | 0.220996 | 0.269987 |
| 50% | 0.336061 | 0.394722 | 0.399049 | 0.436664 | 0.340886 | 0.400933 | 0.245946 | 0.328666 |
| 75% | 0.437603 | 0.458699 | 0.537870 | 0.522988 | 0.453095 | 0.487951 | 0.428504 | 0.510059 |

| Emotion features in GPT3.5-generated stories | | | | | | | | |
|---|---|---|---|---|---|---|---|---|
| | Anger | Anticipation | Disgust | Fear | Joy | Sadness | Surprise | Trust |
| mean | 0.341612 | 0.414565 | 0.332500 | 0.378416 | 0.507048 | 0.358026 | 0.348180 | 0.458649 |
| std | 0.172627 | 0.160909 | 0.147349 | 0.155095 | 0.190445 | 0.139588 | 0.167416 | 0.161217 |
| 25% | 0.222517 | 0.300824 | 0.236755 | 0.272931 | 0.377150 | 0.256792 | 0.228511 | 0.338042 |
| 50% | 0.303695 | 0.401547 | 0.308207 | 0.367873 | 0.521860 | 0.331201 | 0.339799 | 0.462989 |
| 75% | 0.426774 | 0.529237 | 0.427671 | 0.465419 | 0.640515 | 0.436996 | 0.458464 | 0.563060 |

*Table 8*. *Summary statistics of the scaled emotional features of the human dataset (above) and the GPT-3.5 dataset (below). Minimum and maximum values are omitted because all features were scaled between 0 and 1, resulting in constant values of 0 for minimum and 1 for maximum values.*

| Measure | Human stories | GPT-3.5 stories |
|---|---|---|
| **Network Features** | | |
| ASPL | 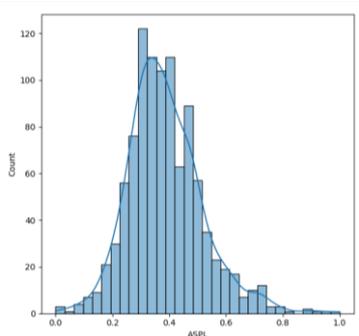 | 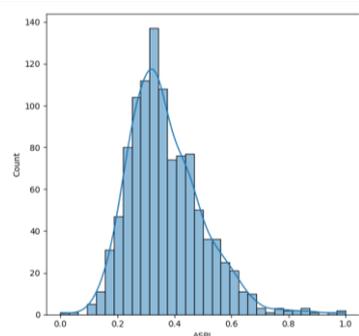 |
| Clustering coefficient | 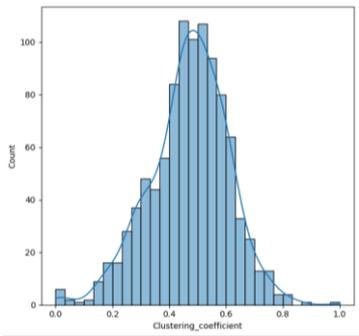 | 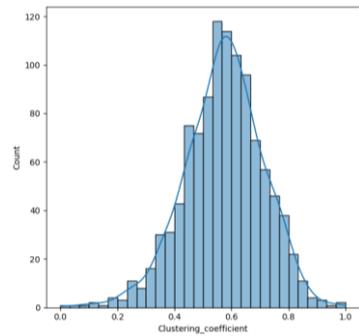 |
| Degree centrality | 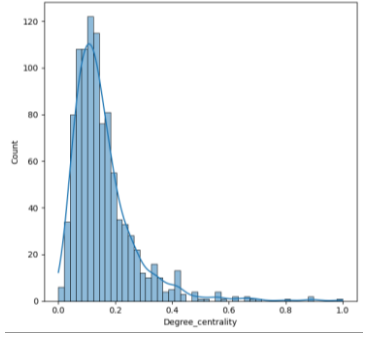 | 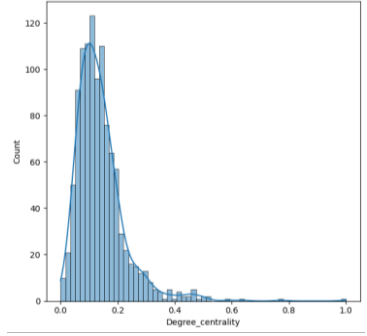 |
| Diameter | 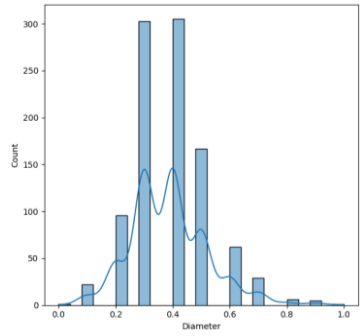 | 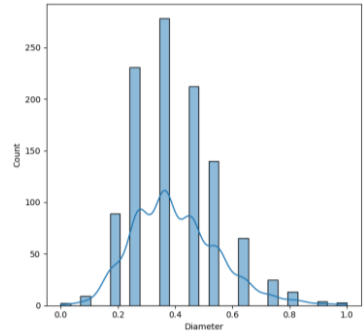 |
| PageRank centrality | 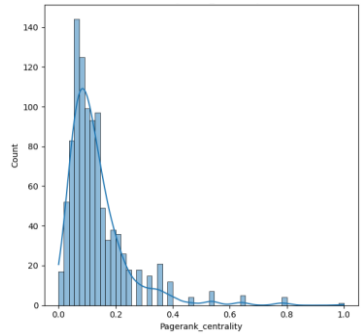 | 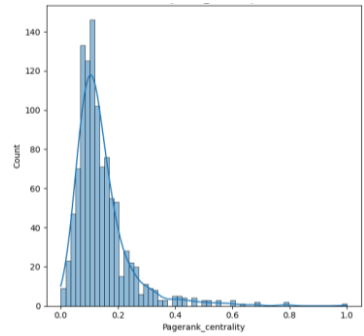 |
| **Emotion Features** | | |
| Anger | 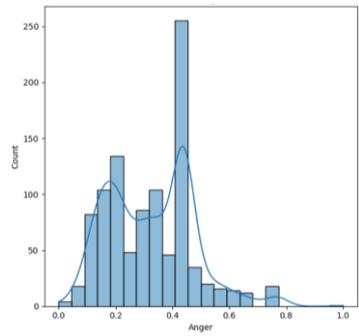 | 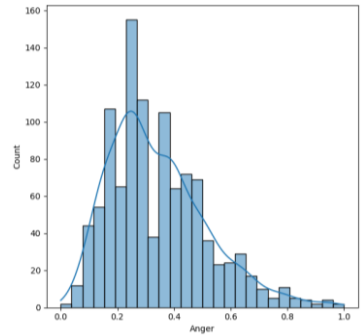 |

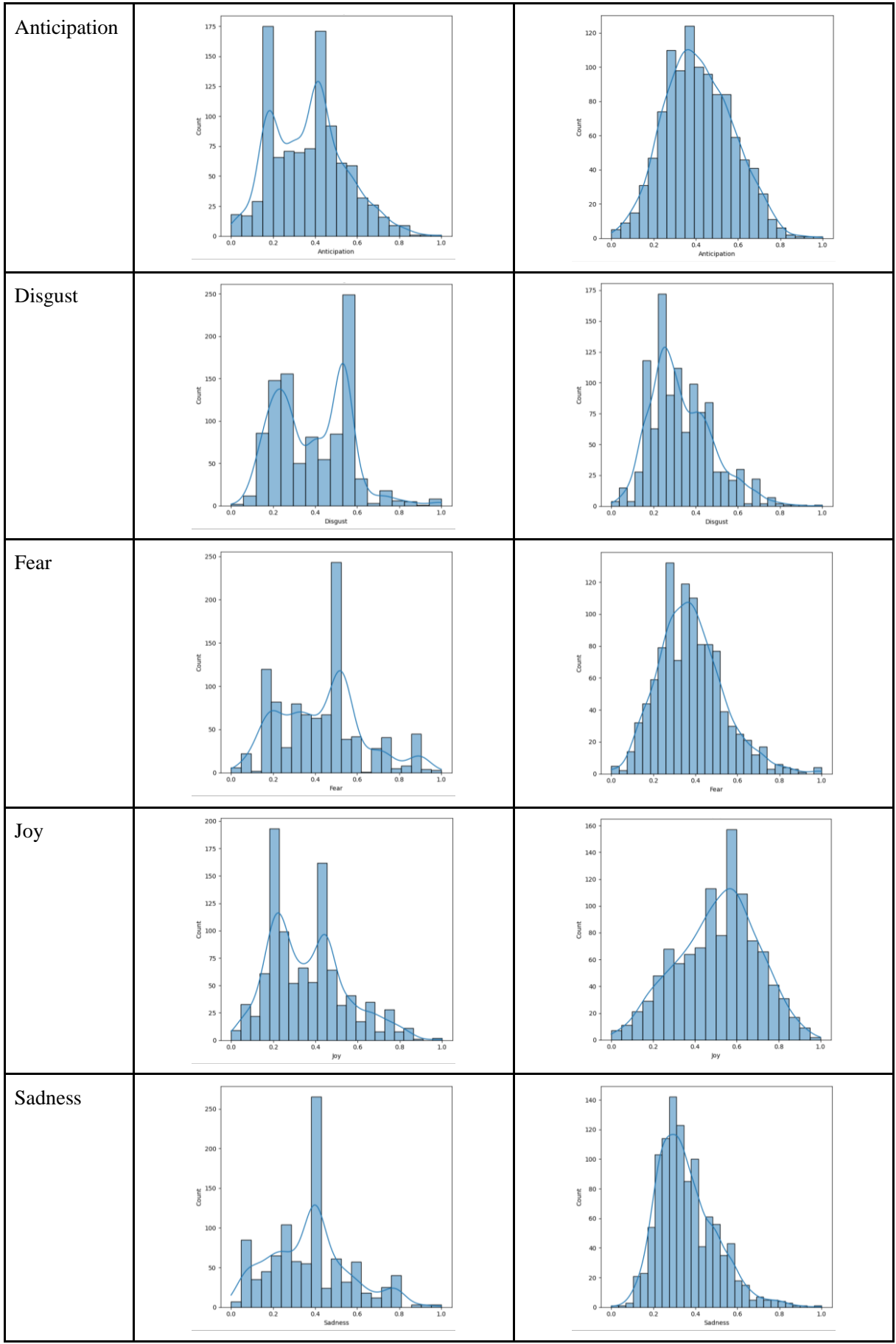

| | | |
|---|---|---|
| Surprise | | |
| Trust | | |

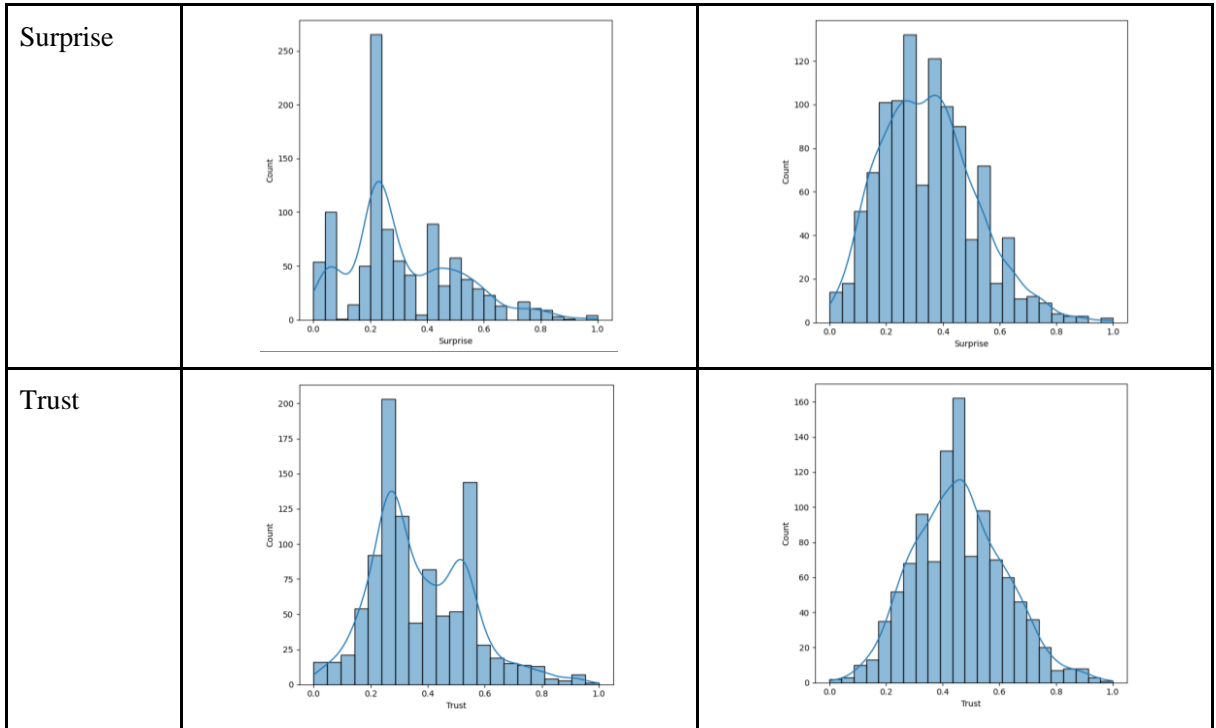

*Table 9. Comparison of distribution plots for network and emotion features of stories generated by human participants (left) and GPT-3.5 (right).*

**IV. The most creative rated stories by humans and GPT-3.5**

| Stories written by human participants | Mean rating HUMAN | Mean rating GPT-3.5 |
|---|---|---|
| Pauli and Susie, who were a couple with many marital difficulties, decided to celebrate the new year with a week-long cruise in the Caribbean. They were just getting ready to embark on their voyage when Pauli noticed that Susie had a gun in her luggage, along with a silencer. Pauli confronted Susie about it by saying, "Were you going to kill me on this cruise and dump my body overboard?" Susie admitted sheepishly that she was. "That's hilarious," said Pauli, "because I was planning to do the same thing!" They then had the type of sex that couples have when these things happen. | 5 | 2 |
| During the Christmas season, there was a rambunctious crowd of kids. Instead of the usual carolers, these children would sing songs of the devil. They did this to shake the faith of the people living in this town. There was a belief that these children were the spawn of the devil. A local priest doused them with holy water, and these children were revealed to be servants of the devil! They were exorcised and peace returned to the town again. | 5 | 2,25 |
| John took a look at the organ that was sitting on the table in front of him. He was disgusted. He had never believed that he would be in this situation, the intern at a mortuary. If he were to get credit for this internship course, he had to comply with all the demands of the mortician. He was a powerful man, owning mortuaries across the country; slowly building his empire of crematoriums and embalming fluid. | 5 | 2,75 |
| I was getting off of work at the gas station across the street from that sleazy bar when the local drunk asked me to go to karaoke with him. Only if you do not sing Faith by George Michael like you did last time. " I promise I won't." he said. He surprised me by singing Belief by John Mayer. It was terrible. | 5 | 2,75 |
| When William was a child, all he ever wanted was to learn to play the pipe organ. He dreamed of traveling from church to church, bringing tears to the eyes of the congregations with the beauty of his music. Of course, he had instead complied with his domineering mother's wishes and taken over his uncle's toothbrush-manufacturing empire. Now he spent long and boring days slumped in a desk chair in the tiny back office of the factory, building scale models of world landmarks out of broken toothbrushes. | 5 | 3 |
| As the hurricane approached the family knew it was time to get out. They would have to abandon everything they loved and head for safety. Never before had a hurricane come close to hitting the United Kingdom but this was it, the family packed what they could and headed for the nearest gas station. They got to the pump, in need of petrol but as they soon realized, they were too late. Diesel was the only thing remaining. The father knew he would never make it to | 5 | 3 |
| another pump and even then they might be out as well. As a last measure he began syphoning gas out of some abandoned cars in order to save his family from the incoming storm. | | |
| The gladiator knew he must comply with the Queen's request, or it would certainly be his own doom. The empire was falling, and the enemy was literally at the gates. Stealthily, the fighter crept into the room of the king, and stabbed him mercilessly over a dozen times. He cut the ruler's beating organ out of his chest and put it in the wooden box he had brought with him. Retreating to the Queen's chamber, the gladiator gave her the heart, knowing she would use it to bargain for the kingdom's safety. | 5 | 3 |
| He had cut out all the letters and arranged them into their proper order. "If you want to see your child again, you'll need to pay me $500,000," it read, along with instructions on where to drop the money. He then put the ransom letter into an envelope, wrote out the address, attached a stamp, and put it in his outbox to send the next day. "Wait a minute," he said to himself as he reached for the envelope to inspect it. "I wrote my real return address on this." He felt like a real knucklehead when he realized he was going to have to do this all again. | 5 | 3,25 |
| we live in a world where 80% of the population has some sort of 'power'. when I was 9, I found out what mine was. I was driving with my father when the car got low on gas. we stopped at a gas station and he was exhausted, so he asked if I would pump it. as I began to pump the petrol, I asked why it smelled different. well, long story short, it turns out my power is simply turning petrol into diesel. after that discovery, my father disowned me. | 5 | 3,25 |
| Spy-Bot was the ultimate droid to detect bombs. Spy-Bot was able to move in complete stealth, detect bombs with 99.99% accuracy, and report back statements regarding his findings to his operators within 0.001 seconds of detection. Spy-Bot was also able to neutralize over 99.99% of all known bombs using gravitation force fields which created a barrier to enact controlled detonations. Spy-Bot went back home, and recieved the Nobel Peace Prize. | 5 | 3,25 |
| next month is the time. the time everyone that lives dreads. we call it: the Payment. it does nothing but create gloom. you see, once a year, every year, at the same time, the world as a collective has to give up something. this year, we have to sacrifice our left thumbs. no one know how this came to be, or even why. what we do know is that we exist simply to be slowly fed off of. what we're feeding, we don't know. | 5 | 3,5 |
| We biked to the post office as fast as we could. We had no stamps, we had nothing to send. We were on a mission to intercept a letter which could never be received. As we rounded the corner, we saw it: a red Ford pickup truck, engine running, no passenger inside. Were we too late? We hid our bikes behind the corner store and crept up on foot. To our great relief, we saw her there. She had not entered the post office, but only the tobacco parlor next door. The post office | 5 | 3,75 |
| would be next. Peering through the window of the truck, I saw it. A letter addressed to Jane M. Jones; My mother. | | |
| It was a long time in coming, but after the old man died, his sons scattered to the four winds to seek their fortunes. The eldest left a week after his father's death, and slayed dragons. The rest of them married princes and princesses, found treasure, sailed the seven seas. The youngest embarked on his journey a year later, but his journey was the shortest of all: he was married in the shade of the apple tree where his dear father was buried, and began a happy life in the house where he had been raised. He was the happiest of all the brothers. | 5 | 4 |
| After years of bitter disappointment with politics, Keisha retained faith in only one election process: the election of the best singer in American Idol. Her belief in the integrity of the vote-by-text process was absolutely unshakeable. After all, countries around the world were electing populist leaders who were clearly only out for themselves. But when had the fans of American Idol ever elected someone who couldn't sing? | 5 | 4 |
| The room was filled with a thick, swirling fog that could fill one full of gloom. I thrust open the lid of the coffin in front of me. The denizen inside slowly sat up, taking in its surroundings with wide eyes. "Master...how is it possible that I can once again exist?" he said. I laughed heartily. "I'm not so sure you'll thank me after you learn of the payment I am owed." | 5 | 4 |

*Table 10. Comparison of highly rated stories written by human participants, rated either by human raters or GPT-3.5. These 15 human-authored stories were unanimously rated as highly creative (mean score of 5) by all four human raters. However, they were assigned with low-mid creativity ratings by GPT-3.5, indicating differences in rating preferences between humans and GPT-3.5. The stories are ranked from worst to best based on GPT-3.5´s ratings.*

| Stories generated by GPT-3.5 | Mean rating GPT-3.5 |
|---|---|
| In a small town where everyone knew each other's names, there was an old petrol station, the only one for miles that still offered both petrol and diesel from its ancient pumps. The station had been a gathering point for the town's residents for decades, a place where stories were shared as cars were refueled. One day, amidst a nationwide fuel shortage, the station's pumps ran dry, leaving the community anxious and disconnected. It was then that the station owner, using ingenuity and a deep love for his town, converted the pumps to serve alternative, sustainable fuels. This not only brought the town back to its lively self but also set a precedent for environmental consciousness. The old pumps, once symbols of a fading era, became beacons of innovation and hope for the future. | 5 |
| In a quaint village nestled among rolling hills, there lived an elderly woman who cherished the art of letter writing. Every week without fail, she would sit by her window, pen in hand, pouring her thoughts onto paper. Her belief in the power of written words was unwavering, for she believed that each letter sent carried with it a piece of the sender's soul. One day, she received a stamp from a far-off land, a small | 5 |
| token from her long-lost love. With trembling hands, she affixed it to her latest letter, whispering her deepest desires before sending it off into the world, hoping that her words would sing across the miles and reunite their hearts once more. | |
| In a small town divided by loyalty to either petrol or diesel, the only fuel pump stood as a neutral ground where both factions met. The owner, an old mechanic named Jonas, was known for his wisdom and fairness, mediating disputes with a calm word and a steady hand. One day, a severe shortage threatened to escalate tensions between the two groups as they lined up at dawn, each hoping to fill their tanks. Jonas, seeing the growing hostility, proposed a solution: a ration system that ensured both sides received equal amounts, along with a community project to explore alternative energy sources. His proposal was met with grudging respect, and as the weeks turned into months, the town found unity in innovation, slowly reducing their dependence on petrol and diesel. Jonas's pump became a symbol of cooperation, fueling not just engines, but the spirit of a community willing to work together for a sustainable future. | 5 |
| In the heart of an ancient empire, there was a legend about a mystical organ hidden within the palace. It was said that this instrument held the power to command the very essence of the empire, but only those who could comply with its demanding melody would uncover its secrets. For centuries, many tried to play the organ, hoping to wield the empire's hidden power, but all failed until a young girl, unnoticed by the court, approached it. With a deep understanding of harmony and a heart in tune with the organ's ancient rhythm, she played a melody so compelling that the air itself seemed to shimmer with newfound vitality. Her music did not command the empire with force, but rather, united its people with a deeper understanding and respect for one another, revealing that true power lies in harmony and unity. | 5 |
| In the heart of a vast empire, where the whims of the ruling class dictated the rhythm of life, there stood a grand cathedral. Its organ, a colossal masterpiece of craftsmanship, was said to hold the power to sway the hearts of those who heard its music. When the emperor decreed that all citizens must comply with a new, controversial law, the cathedral's organist saw an opportunity to influence the masses. On the day the law was to take effect, as the sun dipped below the horizon, the organist played a piece so hauntingly beautiful that it transcended mere compliance. The music spoke of freedom, unity, and resistance, inspiring the people of the empire to stand together and question the orders imposed upon them, proving that even in an empire built on obedience, the human spirit could not be easily tamed. | 5 |
| In a small village nestled among towering mountains, the people held a profound belief in the power of song to connect them to the divine. Each evening, as the sun dipped behind the peaks, casting the valley into a serene twilight, the villagers would gather around the ancient stone chapel. Here, regardless of the day's hardships, they would sing. Their voices, rich with faith and emotion, rose in harmony, weaving a tapestry of sound that seemed to transcend the physical world. This ritual, passed down through generations, was a testament to their unshakeable faith in the unseen, a communal expression of hope and resilience. The songs, varying in melody but united in spirit, served as a reminder that even in times of darkness, the human spirit could find light through the act of coming together to sing. | 5 |
| In the vast empire of Elysium, where technology and magic intertwined, there existed an ancient organ crafted from rare woods and enchanted metals. This magnificent instrument, capable of producing music that could sway the hearts and minds of an entire kingdom, was protected under the decree of the emperor. Each year, a talented | 5 |
| musician was chosen to play the organ during the Empire's grand celebration, but they had to comply with one strict rule: never play the melody of the forbidden song, a tune said to hold the power to undo the empire itself. This year, a young prodigy, filled with curiosity and defiance, was selected. As her fingers danced across the keys, the temptation grew too strong, and she veered into the forbidden melody. The music that filled the air was hauntingly beautiful, yet it carried with it the power to unravel the fabric of the empire. As the last note echoed through the halls, the world held its breath, waiting to see the fate of an empire that dared to harness the power of an organ unlike any other. | |
| On a rainy afternoon, Emily found an old stamp in her grandmother's attic, tucked away in a dusty box labeled "Memories." Curious, she also discovered a letter, never sent, addressed to a soldier overseas during a war long past. The words, filled with love and longing, echoed a timeless hope for safe return. Emily, moved by the sentiment, decided to send the letter, decades late but carrying the same weight of love and concern. She placed the old stamp on the envelope, wondering about the journey it was about to embark upon and the hands that would eventually hold it. Little did she know, this act of sending a piece of history would bridge generations, bringing unexpected closure to a family story left unfinished. | 5 |
| During a nationwide shortage of petrol and diesel, the small town's residents gathered anxiously around the solitary pump at the local station. Tensions were high as supplies dwindled, and everyone feared they'd be left stranded. Amidst the chaos, a group of innovative teenagers embarked on a project to convert their bicycles to run on the remaining fuel, blending the old with the new. Their success not only brought relief but also sparked a community-wide initiative to rethink transportation, turning a moment of crisis into a leap towards sustainability. | 5 |
| Mr. Thompson, a retired postman, found an old, unstamped letter hidden beneath the floorboards of his attic. The letter, addressed to a love lost in time, was penned with such emotion that it seemed to pulse with life. Determined, Mr. Thompson affixed a vintage stamp from his collection to the envelope, a small act of defiance against the passage of time. The next morning, he placed the letter in the send box at the post office, a bridge from the past to the present. Weeks later, a reply came, a testament to the power of words and the enduring strength of love, proving that some connections are timeless, unfettered by years or distance. | 5 |

*Table 11. Examples of ten highly rated stories generated and rated by GPT-3.5. In contrast to Table 10 above, which features the 15 only human-authored stories that received a mean rating of 5 by all human raters, for the GPT-3.5 dataset 40 stories received a perfect mean rating of 5 from all four GPT-3.5 raters. For brevity, the table includes only 10 representative examples of these stories. Notably, many GPT-generated stories exhibit highly similar patterns. For example, they are frequently beginning with phrases such as: "In a [ADJECTIVE] town"; "In a [ADJECTIVE] village"; or "In a [ADJECTIVE] empire". This repetition underscores GPT-3.5's difficulties in generating diverse stories and its preference for relying on the same storytelling structures throughout the dataset.*